\documentclass[preprint,12pt]{elsarticle}

\usepackage{amssymb}

\usepackage{mathtools}
\usepackage{algorithm}
\usepackage{algpseudocode}
\algnewcommand{\LineComment}[1]{\State \(\triangleright\) #1}
\DeclareMathOperator*{\argmin}{\arg\!\min}
\usepackage{xcolor}
\usepackage{multirow}
\usepackage{booktabs}
\usepackage{subfig}
\usepackage[colorlinks]{hyperref}
\usepackage{tabularx}
\usepackage{caption}

\begin{document}

\begin{frontmatter}



\title{Short-term Wind Speed Forecasting for Power Integration in Smart Grids based on Hybrid LSSVM-SVMD Method}


\author[a1]{Ephrem A. Yekun \corref{cor1}}
\ead{ephrem.admasu@mu.edu.et}
\author[a2]{Alem H. Fitwi}
\author[a3]{Selvi Karpaga Subramanian}
\author[a3]{Anubhav Kumar}
\author[a4]{Teshome Goa Tella}

\cortext[cor1]{Corresponding author: Ephrem A. Yekun}

\affiliation[a1]{organization={School of Electrical and Computer Engineering, Mekelle University},
            country={Ethiopia}}
\affiliation[a2]{organization={Binghamton University},
            city={New York},
            state={},
            country={USA}}
\affiliation[a3]{organization={School of Computer Science Engineering, Galgotias University},
            city={},
            state={Utterpradesh},
            country={India}}
\affiliation[a3]{organization={School of Computer Science Engineering, Galgotias University},
            city={},
            state={Utterpradesh},
            country={India}}
\affiliation[a4]{organization={School of Electrical and Computer Engineering, Addis Ababa Institute of Technology},
            city={Addis Ababa},
            state={Addis Ababa},
            country={Ethiopia}}

\begin{abstract}
Owing to its minimal pollution and efficient energy use, wind energy has become one of the most widely exploited renewable energy resources. The successful integration of wind power into the grid system is contingent upon accurate wind speed forecasting models. However, the task of wind speed forecasting is challenging due to the inherent intermittent characteristics of wind speed. In this paper, a hybrid machine learning approach is developed for predicting short-term wind speed. First, the wind data was decomposed into modal components using Successive Variational Mode Decomposition (SVMD). Then, each sub-signal was fitted into a Least Squares Support Vector Machines (LSSVM) model, with its hyperparameter optimized by a novel variant of Quantum-behaved Particle Swarm Optimization (QPSO), QPSO with elitist breeding (EBQPSO). Second, the residuals making up for the differences between the original wind series and the aggregate of the SVMD modes were modeled using long short-term model (LSTM). Then, the overall predicted values were computed using the aggregate of the LSSVM and the LSTM models. Finally, the performance of the proposed model was compared against state-of-the-art benchmark models for forecasting wind speed using two separate data sets collected from a local wind farm. Empirical results show significant improvement in performance by the proposed method, achieving a 1.21\% to 32.76\% reduction in root mean square error (RMSE) and a 2.05\% to 40.75\% reduction in mean average error (MAE) compared to the benchmark methods. The entire code implementation of this work is freely available in Github\footnote{\url{https://github.com/ephrem-admasu/windspeed_prediction}}.
\end{abstract}

%


\end{frontmatter}


\section{Introduction}
\label{sec:introduction}

The utilization of renewable energy sources has been gaining tremendous attention in recent years, mainly because of the continuing depletion of oil and widespread awareness of climate change \cite{ahuja2016challenges}. Wind energy is preferred over other energy sources for its efficiency, low pollution, and economic profitability. Wind turbines, devices used in converting wind's kinetic energy of wind into electrical energy, are cheaper to install and operate reliably without incurring too much cost. Moreover, they can function for more than 8 years, and their decommissioning process is environmentally friendly. Hence, the application of wind speed for energy harvesting has become the fastest growing endeavor in the last few years \cite{liu2020short}. The promotion of electrification using wind power has been increasing, and many wind farms are being deployed around the world. In 2022, for example, 906 gigawatts (GW) of total installed capacity were reached \cite{globalwind2023} and this trend is expected to keep soaring in the decades ahead of us.

Although wind power generation has many benefits, it also comes with nontrivial challenges that power system operators face when they integrate wind energy into the grid system. Particularly, the non-stationary, limited dispatchability, and non-storability of wind are the main characteristics that make integration strenuous and challenging. Wind energy has limited dispatchability in the sense that it cannot be increased when needed; it can only be reduced. Also, wind cannot be stored, as in the case of atoms and coal, for future use \cite{zhu2012short}. The goal of power system operators is to maintain balance between the power supply and electric demand under uncertainties and transmission network constraints without causing damage or incurring heavy costs. Therefore, power system operators come to grips with the situation of balancing energy demand and supply while operating under the limiting characteristics of wind energy.

By predicting the wind power and the load demand ahead of time, balance can be maintained with little cost. However, the operation cost and system instability are higher when large-scale wind power is integrated into the electric grid, mainly because wind speed is intermittent and difficult to predict. When less power is produced due to inaccurate predictions, the damage can be huge, ranging from blackouts to resorting to costly operations such as gas-fired plants to maintain balance \cite{xie2010wind}. Therefore, predicting wind speed accurately and robustly is crucial to integrating wind power into the grid with minimum costs.

The ultimate task of wind power forecasting is achieved by utilizing two techniques. The first strategy includes forecasting wind speed first and then calculating the wind power generated by each turbine using a deterministic power curve that is supplied by the turbine manufacturer. The second strategy involves forecasting wind power directly. The first method is mostly preferred for power system operations for three main reasons. First, two nearby wind farms running different types of wind turbines may receive the same wind speed. Forecasting wind power for each farm separately may be impractical when it can easily be achieved using the common predicted wind speed and the power curves of each turbine. Second, wind turbines manufactured by different manufacturers can be erected in a single wind farm. This is the case of the Ashegoda wind farm, which is located near Mekelle City, Ethiopia, and from which we collected our dataset and on which our study is based. The wind farm operates using two types of wind turbines, manufactured by a Spanish company and a French company. Finally, owing to the spatial correlation of wind, forecasting wind speed can be more accurate than forecasting wind power. This becomes more noticeable in the absence of wind farms or other sources of wind energy, where forecasting upstream wind speeds may be made in order to predict the wind power production of a wind farm located downstream of the wind \cite{zhu2012short}.

Wind forecasting horizon is generally grouped into four categories: ultrashort-term forecasting, which predicts wind within a few seconds to 30 minutes; short-term forecasting, which conducts prediction between 30 minutes and 6 hours ahead; medium-term prediction, which predicts wind between 6 hours and one day; and long-term forecasting, in which the forecast is beyond one day ahead \cite{xie2022overview}. Ultrashort-term and short-term wind forecasting are sometimes considered one, and their applications mainly involve power system operation security and reliability \cite{liu2021application} such as real-time grid operation, economic load dispatch planning, load reasonable decisions, and operational security in the electricity market \cite{meenal2022weather}. Therefore, this work mainly focuses on short-term prediction of wind speed for wind farms.

Nowadays, many nations have reached an agreement on the importance of green, low-carbon, and sustainable development and the pursuit of carbon neutrality goals. For instance, China offered the target of "peak carbon dioxide emissions by 2030 and carbon neutrality by 2060," whereas Finland and Europe both put out the target of 2035 \cite{sheng2023short}. The energy sector as a whole produces a significant amount of carbon emissions. To achieve carbon neutrality in the energy sector, the power system must first take the lead, and wind power is the key enabler \cite{sun2022multi}, \cite{jalali2021towards}.

The main focus of this work is to propose a novel AI-based hybrid model for short-term wind speed forecasting (STWSF) using data collected from a local wind farm. An SVMD decomposition scheme is used to create intrinsic modes, where each mode is fed into an LSSVM model. LSSVM is an efficient algorithm that is used in regression applications, including WSF. The performance of LSSVM is heavily affected by two of its hyperparameters, known as the regularization parameter and the kernel parameter, which have to be tuned carefully. To that end, a new variant of QPSO, known as QPSO with elitist breeding (EBQPSO), is used to select the optimum values of the two hyper-parameters during the training and validation stages for each mode generated by SVMD. Further, the trend of the error sequence obtained from sub-signal modes and the original wind data is modeled by LSTM in order to further improve accuracy and maintain stable performance. Finally, an aggregate of the predicted intrinsic modes and error sequence is taken to produce the final prediction output wind speed.

The remaining content of this paper is organized as follows: in \hyperlink{sec:review}{Section 2}, a close study of related works is provided. \hyperlink{sec:method}{Section 3} presents a detailed discussion of the methodology followed. This includes the theoretical and mathematical foundations of SVMD, LSSVM, and the proposed improved QPSO algorithm and outlines the comparative advantage of the algorithm as compared to PSO and QPSO. This section discusses wind data acquisition and preprocesing, feature extraction and selection from the preprocessed data, and hyperparameter breeding procedures for LSSVM based on the proposed optimization method. Moreover, it outlines the procedures followed for training and optimization of the overall STWSF model. Section \hyperlink{sec:results}{Section 4} discusses the experimental results of the proposed model and compares its performance to other known models used in STWSF using various performance metrics. Finally, \hyperlink{sec:conc}{Section 5} summarizes the concluding remarks, and recommends further research work. 

\section{Literature Review}
\label{sec:literature}
The task of accurately forecasting wind speed and wind power is in high demand. Many researchers in academia and industry are conducting myriad research projects on wind speed forecasting. In general, wind forecasting techniques are categorized into four categories: physical methods, statistical methods, AI-based methods, and hybrid or combined methods \cite{liu2020combined}. Physical methods work according to numerical weather prediction (NWP) and utilize weather variables such as temperature, pressure, and surface roughness. Using a proposed sequence transfer correction algorithm (STCA) for NWP wind speed sequence, Wang et al. \cite{wang2019sequence} developed a novel forecasting method. Groch et al. leveraged a mean-variance portfolio optimization technique and a two-layered feed-forward neural network to implement a Weather Research and Forecasting (WRF) model and enhanced the overall performance with the ensemble members \cite{groch2019short}. Higashiyamaa et al. \cite{higashiyama2018feature} developed a three-dimensional convolutional neural network (3D-CNN) to extract spatiotemporal entities from NWP data automatically. NWP solves complicated and chaotic differential equations by utilizing high-computing resources from supercomputers, which may last a day or more. For this reason, these techniques are not feasible for short-term forecasting \cite{wang2019sequence}, causing the research community to look for other solutions.

Statistical models, also known as time-series methods, forecast wind speed by exploiting historical data. The most widely used techniques are called Autoregressive Moving Average (ARMA) \cite{erdem2011arma}, \cite{yunus2015arima}, \cite{cadenas2016wind}, \cite{lydia2016linear}. Using an ARMA method, Erdem \cite{erdem2011arma} performed wind speed and direction forecasting, first by separating the wind speed into lateral and longitudinal spaces, forecasting each individually, and combining them to obtain the final forecasts. One-step wind speed forecasting was performed for two different regions in \cite{cadenas2016wind}. The authors compared two models for both regions: a linear autoregressive integrated moving average (ARIMA) and a nonlinear autoregressive exogenous artificial neural network (NARX) model, which produced similar results. After dividing the wind speed data into its frequency components, an ARIMA method was used in \cite{yunus2015arima}. The authors also demonstrated the model's applicability to estimate wind speed for neighboring areas. A 10 minutes ahead WSF model was developed by Lidya et al. using a nonlinear  ARMA models with eXogenous (ARMAX) \cite{lydia2016linear}. 
The ARMAX technique for wind direction produced superior performance as compared to other ARMAX modles. Most of the statistical models have simple structures, fast computation time, and strong interpretation capability; however, with more time, the forecasting accuracy decreases. Since the relationship between the input and output is maintained with precise mathematical equations, there is an issue of adaptability for prediction. Statistical methods also fail to capture nonlinear relationships in the time series \cite{zhang2020short}. Overall, statistical methods are more suitable for stationary time series than non-stationary series, and real-world wind data are mostly non-stationary.

AI-based algorithms are methods that utilize machine learning and deep learning algorithms. Their good accuracy and strong capability of handling nonlinear and non-stationary data make them preferable for wind forecasting applications. Examples include artificial neural networks (ANN) \cite{li2010comparing, bhaskar2012awnn, wu2015short}, extreme learning machine (ELM) \cite{luo2018short, sun2018adaptive, sun2018adaptive}, support vector machine \cite{zhou2011fine, yang2015support}, least squares support vector machine (LSSVM) \cite{hu2014short, xiang2019forecasting}, and fuzzy logic \cite{an2014fuzzy, khorramdel2018fuzzy}. Due to their superb capacity for self-learning and non-linear mapping, ANNs have been widely used for WSF applications. However, ANN models can easily be trapped into local minima during training since defining crucial parameters such as learning rate, number of iterations, and trapping criteria is usually arduous. As an alternative, ELM models have been applied for WSF and produced good results. But since ELM models only take into account the empirical risk minimization principle at the training phase, their overall performance may be unreliable \cite{sun2018adaptive}.

LSSVM algorithm is an improved version of SVM that deals with linear equations rather than the quadratic formulations of SVM. Despite the widespread use of this method in WSF, LSSVM relies on two important hyperparameters that greatly affect the overall prediction performance. These two hyperparameters, known as the regularization parameter ($\gamma$) and kernel parameter ($\sigma ^ 2$), have to be chosen carefully in order to avoid overfitting or underfitting. To that end, various optimization techniques have been proposed. In \cite{sun2012short}, a PSO algorithm was proposed to obtain optimized parameters of LSSVM trained using a dataset gathered from a wind farm in Inner Mongolia. A short-term WSF method using an improved PSO method was also used by Chang et. al. \cite{chang2013short} in a combination of the persistence method, a radial basis function (RBF), and a neural network. The proposed method was able to accurately predict the actual values. The PSO method can produce improved forecasting results as compared to using the default parameters of LSSVM. However, PSO is prone to being stuck at local minima during the search procedure. In an effort to solve this problem, a variant of QPSO that allocates weighted values into the particles was proposed by Hu et. al. \cite{hu2014short} and used the method to generate optimum values of LSSVM parameters. Despite the method's overall boost in performance, the authors suggested further improvement can be achieved by intelligently tweaking the weights in a fixed interval \cite{hu2014short}.

Deep learning methods, a subcategory of AI-based approaches, have also been widely used for wind speed forecasting. The most commonly implemented models include recurrent neural networks (RNN) \cite{li2019multi}, convolutional neural networks (CNN) \cite{harbola2019one}, and long-short-term models (LSTM) \cite{geng2020short} among others. In \cite{li2019multi}, a multi-step wind speed prediction framework using a Wind Speed and Turbulence Intensity-based RNN was presented. The proposed scheme displayed superior performance compared to other machine learning methods, but issues related to the relationship between turbulence intensity and the performance of different time intervals were not addressed. Using historical data on wind speed and direction, Harbola et al. proposed a two-layer 1-D CNN to predict future values of wind speed and direction \cite{harbola2019one}. Although good results were obtained, the authors concluded that accuracy can be further improved by increasing the number of feature maps and the number of neurons using more hardware resources. An LSTM model was implemented in \cite{geng2020short} where a principal component analysis (PCA) was first used to reduce the meteorological data dimensions and a differential evolution (DE) algorithm was applied to generate optimized values of LSTM hyperparameters such as the learning rate, number of hidden layer nodes, and batch size. Overall, the advantages of deep learning models become more pronounced when there is a supply of huge amounts of data and a lot of computational resources. However, with relatively average data and computational resources, deep learning models fail to produce more accurate results than LSSVM models \cite{tian2020short}.

Taking into account the intermittence and volatility of wind time series and the limitations of the aforementioned methods, scholars have been striving to develop more robust and accurate models. In recent years, hybrid methods have been the most successful in the task of wind speed prediction as they leverage the strengths of two or more different methods \cite{qin2019hybrid, kosana2022hybrid}. Qin et al. \cite{qin2019hybrid}, for instance, used CNN and LSTM with a multi-task learning strategy to predict short-term wind speed. Another way to create hybrid models is to use time-series processing techniques, particularly decomposition algorithms. With time series decomposition, the wind series is first broken down into sub-series of different frequencies. A forecasting model is then used to predict each subsequence. Finally, the aggregate of the prediction outputs of each sub-sequence is taken to be the ultimate prediction result \cite{li2022multi}.

The decomposition techniques most commonly employed for WSF are empirical mode decomposition (EMD) \cite{bokde2019review}, ensemble empirical mode decomposition (EEMD) \cite{chen2021short}, and variation mode decomposition (VMD) \cite{zhu2022short}. EMD- and EEMD-based decomposition methods have been shown to generate moderate accuracy gains. However, due to the mode aliasing phenomenon that occurs in EMD and the prevalence of high noise in the residue of EEMD, the accuracy gains are limited \cite{song2018novel}. VMD is an alternative decomposition scheme over EMD and EEMD and has produced good results for wind speed applications \cite{zhang2019novel, zhu2022short}. However, the performance of VMD depends on important parameters, such as the number of components, that must be predefined by the user. Failure to choose the correct parameters gives rise to mode mixing or mode splitting in the decomposed components \cite{eriksen2023data} and can greatly affect the overall accuracy of the WSF model. To mitigate these problems, successive VMD (SVMD) was proposed to select parameters in an iterative and automatic manner \cite{nazari2020successive}. Therefore, SVMD is a better alternative for decomposing time series data such as wind speed. 


This study proposes a hybrid model for short-term WSF that takes into account 1) the suitability of LSSVM models for average data sizes and computational resources with an improved QPSO algorithm for optimizing its parameters, 2) the ability of LSTM networks to model irregular sequences, and 3) the advantages of the SVMD decomposition algorithm over other techniques. This work is novel since it is the first to attempt to use an improved QPSO algorithm based on the principle of transposon operators to optimize LSSVM parameters. Moreover, the hybrid SVMD-EBQPSO-LSSVM-LSTM model is the first to be employed for the purpose of short-term wind speed forecasting.

The main contributions of this work are highlighted as follows: \begin{itemize} 
\item The original QPSO optimization algorithm is improved using the concept of elitist breeding where the personal best of each particle and the global best are interbred using a mechanism known as the transposon operator. 
\item Successive variational mode decomposition (SVMD) is introduced to break down the preprocessed wind speed data of two datasets, collected from a local wind farm, into separate modes. We show that the SVMD improves the forecasting accuracy of the proposed model. \item The improved QPSO, known as QPSO with elitist breeding (EBQSPO), is used to search for the for the optimum parameters of the LSSM algorithm and the window size for each mode. 
\item The error sequence making up for the difference between the aggregated SVMD modes and the original wind speed series was built by an LSTM network. The capability of LSTM networks to memorize long historical trends and irregularities makes them ideal for modeling error sequences of the kind presented. The final forecasting results were then computed by combining the aggregate of the predicted modes and the error series. 
\item Finally, a comprehensive performance evaluation of the proposed model against competitive benchmark models, such as SVMD-CNN, SVMD-LSTM, and SVMD-LSSVM, is presented to show the superiority of the proposed model. 
\end{itemize}

\section{Methodology}
\label{sec:methodology}
\subsection{Successive Variational mode decomposition}
Variational mode decomposition (VMD) was first introduced by Dragomiretskiy and Zosso in 2014 \cite{dragomiretskiy2013variational}. It is an adaptive signal decomposition scheme that takes a time-series signal $f(t)$ and decomposes it into $K$ sub-signals (or modes) $u_k \,\, (k \in 1, 2, \dots k)$ in a non-recursive manner. Unlike EMD and its variants, VMD is developed under a well-founded mathematical theory and is not prone to sampling rate and noise \cite{li2017research}. The main issue with VMD is that the number of modes, $K$, must be chosen before the decomposition process starts. Low K may result in duplicate modes, whereas high K may produce mode mixing or just noisy modes \cite{eriksen2023data}. Therefore, choosing the parameter $K$ incorrectly will cause the performance of the algorithm to suffer, resulting in a degraded overall performance of the of the wind speed prediction model.

In an effort to solve this issue, Nazari et al. \cite{nazari2020successive} introduced successive variational mode decomposition (SVMD), which extracts the modes of a signal without knowing the value of $K$ in advance. This is accomplished by enforcing some criteria on the optimization problem of VMD. In particular, the spectrum of the mode of interest and the other modes, including the residual, must not overlap, guaranteeing that the most recent mode retrieved is distinct from earlier modes discovered. The SVMD method first assumes the original signal is decomposed into the $L^{th}$ $u_L(t)$ and residual signal $f_r(t)$:

\begin{align}
    \label{eqn:svmd1}
    f(t) = u_L(t) + f_r(t)
\end{align}

The residual signal $f_r(t)$ is also assumed to be composed of two components: the sum of the previously obtained modes the unprocessed part of the original signal $f(t)$ as shown in equation \ref{eqn:svmd2}:

\begin{align}
    \label{eqn:svmd2}
    f_r(t) = \sum_{i=1}^{L-1} u_i(t) + f_u(t)
\end{align}

To extract the $L^{th}$ mode, the SVMD method enforces the following criteria: First, each mode is expected to be compact around its center frequency, which is also the main criterion used by VMD. Second, there should be minimal or no spectral overlap between $u_L(t)$ and $f_r(t)$. Third, the energy of the residual signal $f_r(t)$ around the frequency components of the $L_{th}$ mode $u_L(t)$ should be minimized. Finally, the original signal should be completely reconstructed from its modes and the residual signal. Taking all these criteria into account, the constrained minimization optimization problem is formulated as in equation \ref{eqn:svmd3}:

\begin{align}
\begin{split}
\label{eqn:svmd3}
\min_{u_L , \, \omega_L , \, f_r} \biggl\{ \alpha \left\| \partial t \Big[ (\delta (t) + \frac{j}{\pi t} ) * u_L(t) \Big] \exp ^ {-j\omega_{k}t} \right\|_2^2 \\
+ \left\| \beta_L(t) * f_r(t) \right\|_2^2 + \sum_{i=1}^{L-1} \left\| \beta_i(t) * u_L(t) \right\|_2^2     \biggl\} \\
\mbox{s.t.} \,\, u_L(t) + f_r(t) = f(t)
\end{split}
\end{align}

In the above equation, $\omega_L$ is the center frequency of the $L^{th}$ mode, $\alpha$ is a balancing parameter, $\delta(t)$ denotes the Dirac distribution, and $*$ is a convolution operation. $\beta(\omega)$ is the impulse response of the filter $\hat{\beta}(\omega)$, which is used to minimize the spectral overlap between the residual and the $L^{th}$ mode signal, and is given as $\hat{\beta}(\omega) = \frac{1}{\alpha(\omega - \omega_L)^2}$ and $\hat{\beta_i}(\omega) = \frac{1}{\alpha(\omega - \omega_i)^2}$. After transforming equation \ref{eqn:svmd3} into an unconstrained optimization problem by introducing the Lagrange multiplier 
$\lambda$, and converting the unconstrained equation into the frequency domain using Parseval's theorem, the final minimization problem is given as follows:

\begin{align}
\begin{split}
\label{eqn:svmd4}
\mathcal{L}(u_L, , \omega_L, , \lambda) = \alpha \left\lVert j(\omega - \omega_L) \big[ (1 + \mbox{sgn}(\omega)) \cdot \hat{u_L}(\omega) \big] \right\rVert|_2^2 \\
+ \left\lVert \hat{\beta_i}(\omega) \cdot (\hat{f_u}(\omega) + \sum_{i=1}^{L-1} \hat{u_i}(\omega)) \right\rVert_2^2  + \sum_{i=1}^{L-1} \left\lVert \hat{\beta_i}(\omega) \cdot \hat{u_L}(\omega) \right\rVert_2^2 \\
+ \left| f(\omega) - \big( \hat{u_L}(\omega) + \hat{f_u}(\omega) + \sum_{i=1}^{L-1} \hat{u_i}(\omega) \big) \right\rVert_2^2
\\
+ \left\langle \hat{\lambda}(\omega), \hat{f}(\omega) - \big( \hat{u_L}(\omega) + \hat{f_u}(\omega) + \sum_{i=1}^{L-1} \hat{u_i}(\omega) \big) \right\rangle
\end{split}
\end{align}

\begin{figure}[!ht]
    \centering
    \includegraphics[scale=.15]{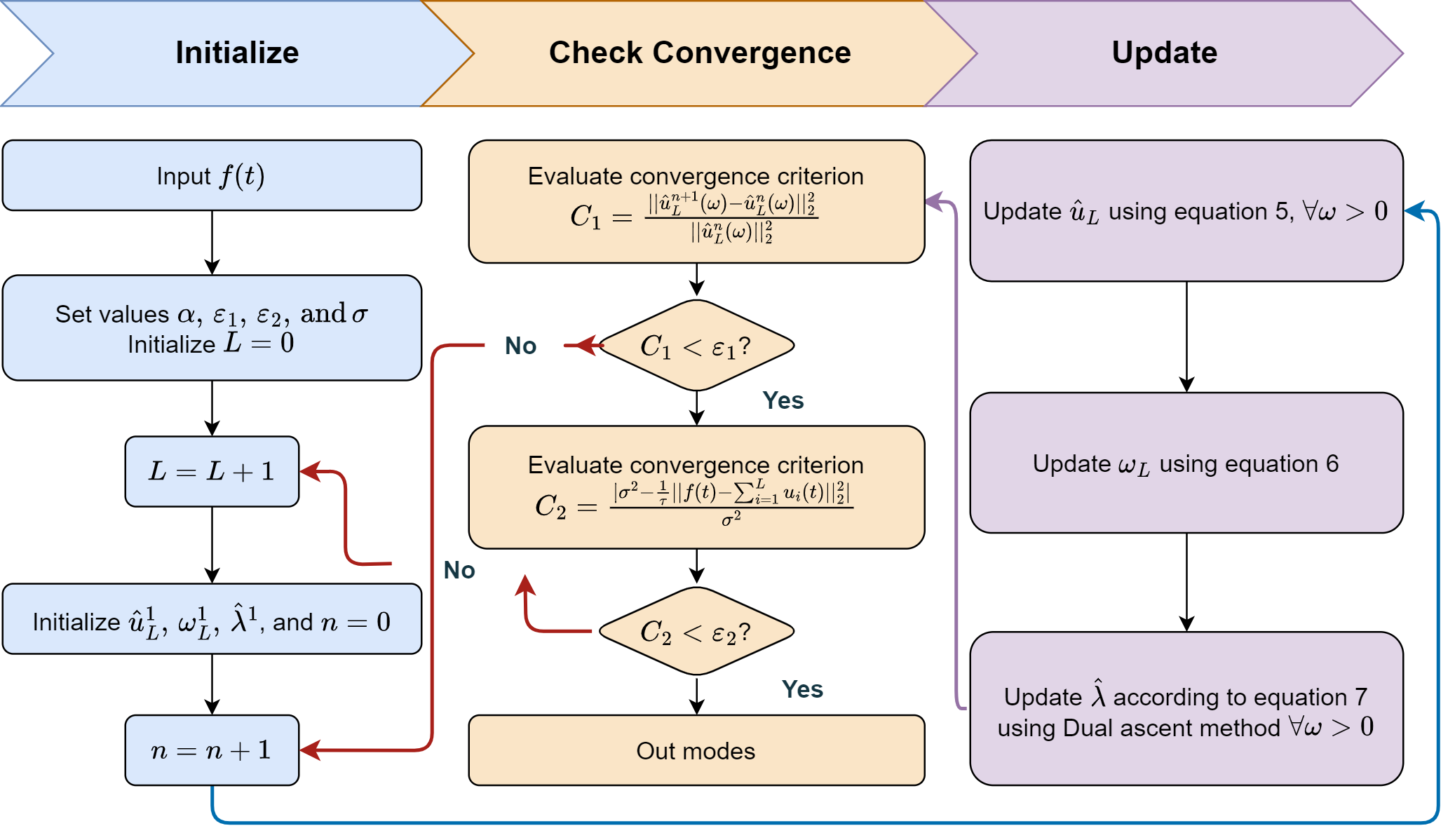}
    \caption{The flowchart for the SVMD method.}
    \label{fig:svmd_flow}
\end{figure}

The solutions of equation \ref{eqn:svmd4} are solved iteratively using the alternate direction method of multipliers (ADMM) as shown in equations \ref{eqn:svmd5} -- \ref{eqn:svmd7}.

\begin{equation}
\label{eqn:svmd5}
 \hat{u}_L^{n+1}(\omega) = \frac{\hat{f}(\omega) + \alpha^2(\omega - \omega_L^n)^4 \cdot \hat{u}_L^n(\omega) + \frac{\hat{\lambda}(\omega)}{2}}{\big[ 1 + \alpha^2(\omega - \omega_L^n)^4 \big] \cdot \Bigg[ 1 + 2\alpha (\omega - \omega_L^n)^2 + \sum_{i=1}^{L-1} \frac{1}{\alpha^2(\omega - \omega_i^n)^4} \Bigg] }
\end{equation}
\begin{equation}
    \label{eqn:svmd6}
    \omega_L^{n+1} = \frac{\int_0^\infty \omega\Big|\hat{u}_L^{n+1}(\omega)\Big|^2 \, d\omega}{\int_0^\infty \Big|\hat{u}_L^{n+1}(\omega)\Big|^2 \, d\omega}
\end{equation}
\begin{multline}
    \label{eqn:svmd7}
    \hat{\lambda}^{n+1}(\omega) = \hat{\lambda}^n(\omega) + \tau \Biggl[ \hat{f}(\omega) - \Biggl( \hat{u}_L^{n+1}(\omega) + \\
   \frac{\alpha^2(\omega - \omega_L^{n+1})^4 \Big( \hat{f}(\omega) - \hat{u}_L{n+1}(\omega) - \sum_{i=1}^{L-1} \hat{u}_i(\omega) \Big) - \sum_{i=1}^{L-1} \hat{u}_i(\omega)}{1 + \alpha^2(\omega - \omega_L^{n+1})^4} \Biggr) \\
  \sum_{i = 1}^{L-1} \hat{u}_i^{n+1}(\omega) \Biggr]   
\end{multline}

where $\hat{f}(\omega)$ and $\hat{u}_L^n$ are the Fourier transforms of the original singal $f(t)$ and the $L^{th}$ mode $u_L^n(t)$ with center frequency $\omega_L^n$ during the $n^{th}$ iteration. $n$ is the current iteration, and $\tau$ is the iteration step length. The overall procedure for the SVMD algorithm is shown in Figure \ref{fig:svmd_flow}.

\begin{figure}
\centering
\includegraphics[scale=.4]{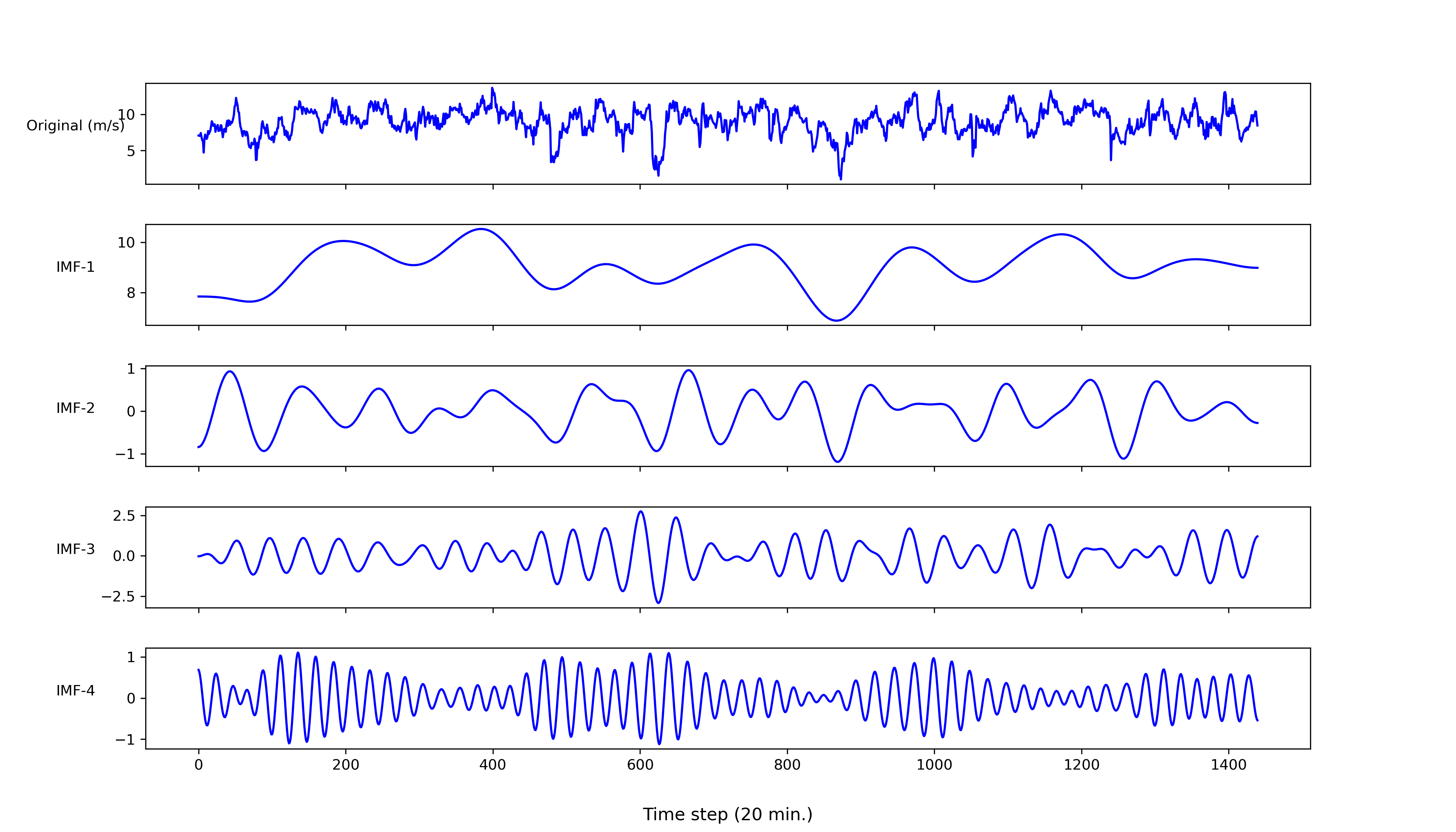}
\caption{\label{fig:vmds1} Original wind speed series and the decomposed modes of SVMD (April Dataset).}
\end{figure}

\begin{figure}
\centering
\includegraphics[scale=.4]{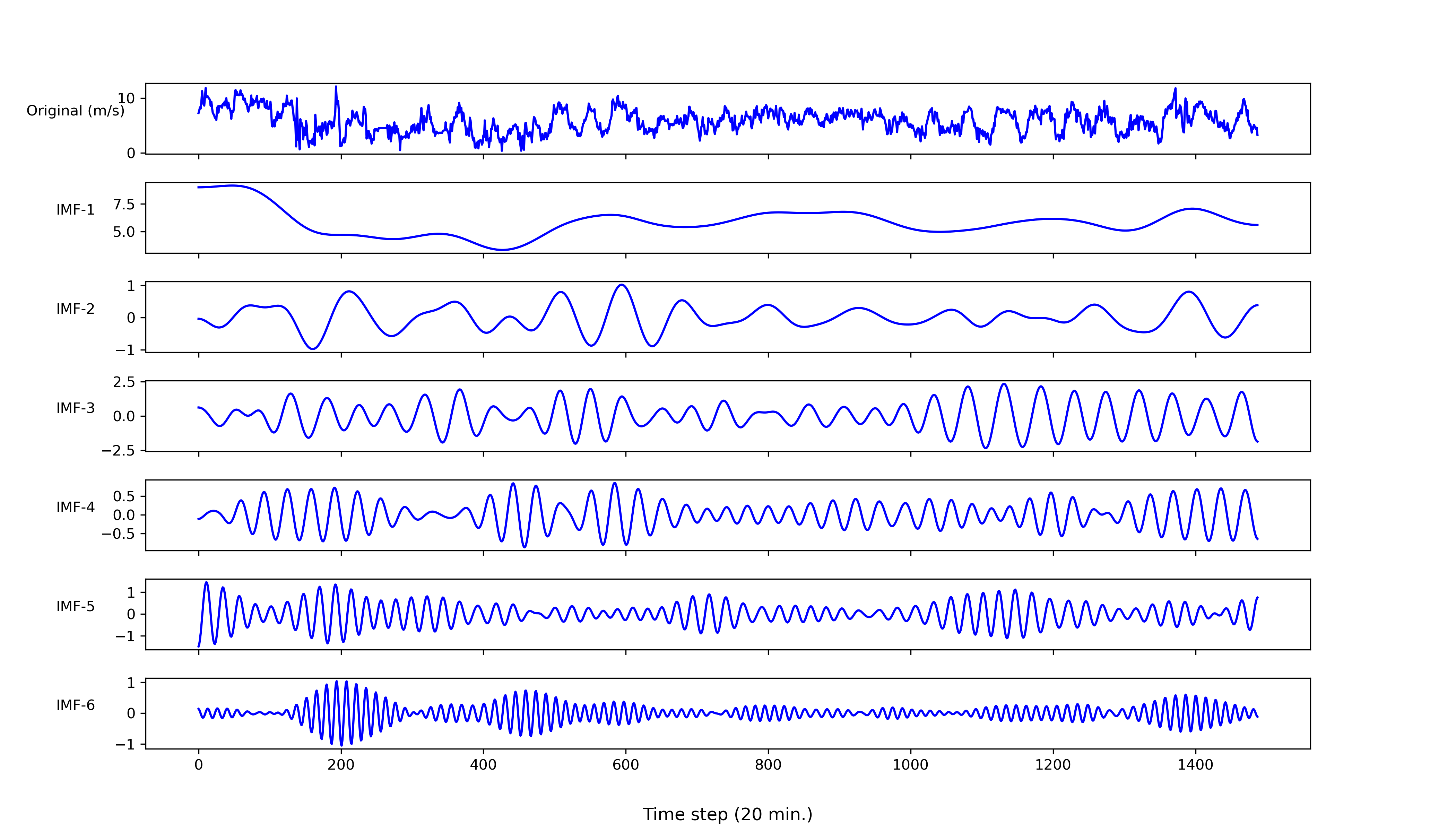}
\caption{\label{fig:vmds2} Original wind speed series and the decomposed modes of SVMD (May Dataset).}
\end{figure}

\subsection{Least squares support vector machines \label{seclssvm}}
LSSVM is an improved version of SVM, which adopts least squares to efficiently distinguish patterns in feature inputs for classification and regression tasks. First proposed by Suykens, LSSVM formulates a linear rather than a quadratic optimization solution, providing an efficiency boost over SVM \cite{suykens1999least}.

The main strength of SVM is its capability to learn sophisticated patterns in a dataset using functions known as kernels and utilizing the sparsity of the final solutions. Nonetheless, the computational complexity of the learning algorithm increases in proportion to the size of the input data in a quadratic manner, making it unfeasible for huge datasets \cite{zhou2011fine}. LSSVM avoids this by converting the inequality constraints into equality constraints, which results in a linear optimization complexity. This also provides a robust loss function and improved overall convergence \cite{hu2014short}.

The LSSVM algorithm works as follows: let the training data be represented as $\{(x_i, y_i), \, x_i \in \mathbb{R} ^ m, y_i \in \mathbb{R}, \, i = 1, \dots, N \}$, where $N$ is the size of the training point, and $m$ is the number of the feature inputs. A linear function is then used to fit the training set based on equation \eqref{eqn:linfun}.
\begin{equation}
\label{eqn:linfun}
f(x) = w^T \phi(x) + b
\end{equation}
where, $\phi(\cdot)$ is a nonlinear mapping function; $w$ is weight vector; $b$ is bias.
Following the risk minimization principle, we can transform equation \eqref{eqn:linfun} into a constrained minimization optimization problem as follows \cite{de2002least}: 
\begin{align}
\begin{split}
\label{eqn:optform}
\min J(w, \xi) = \frac{1}{2}w^Tw + \frac{1}{2}\gamma \sum_{i=1}^N \xi ^2 \\
\mbox{subject to } y_i = w^T\phi(x_i) + b + \xi _i,  \,\, i \in (1, \dots, N)
\end{split}
\end{align}
where, $\gamma$ is the error penalty parameter; $\xi_i$ is the slack variable. Using the Largrange function the constrained optimization problem is transformed into an unconstrained optimization problem as follows:
\begin{equation}
\label{eqn:lagrang}
L = \frac{1}{2}w^Tw + \frac{\gamma}{2} \sum_{i=1}^N \xi^2 - \sum_{i=1}^N a_i(w^T\phi(x_i) + b + \xi _i - y_i)
\end{equation}
$a_i$ is the Lagrange multiplier. After applying the Karush-Kuhn-Tucker (KKT) conditions, we obtain the following linear equation.
\begin{align}
\begin{split}
\label{eqn:kkt}
\left( \begin{array}{cc} 
                0 & \quad \textbf{I}^T \\
                1 & \quad \Omega+\gamma^{-1}\textbf{I} 
        \end{array} 
\right)
\left( \begin{array}{c} 
                        b \\ a 
                \end{array} 
\right)
=
\left( \begin{array}{c} 
                        0 \\ y
                \end{array} 
\right)
\end{split}
\end{align}
Where $y = [y_1, \dots, y_N]^T, \, a = [a_1, \dots, a_N]^T, \, \Omega = \phi(x_i)^T\phi(x_j), \, \mbox{and} \, \textbf{I}$ is a unit matrix.
Mercer condition establishes the kernel function as follows:
\begin{equation}
\label{eqn:kernel}
k(x_i, x_j) = \phi(x_i)^T\phi(x_j)
\end{equation}
We can derive $a$ and $b$ from equations \eqref{eqn:kkt} and \eqref{eqn:kernel} and formulate the final LSSVM regression function using equation \eqref{eqn:lssvm}.
\begin{equation}
\label{eqn:lssvm}
f(x) = \sum_{i=1}^N a_ik(x, x_i) + b
\end{equation}
Several kernel functions are known for LSSVM and SVM. However, the most commonly used kernel function is the Gaussian Radial Basis Function (RBF) kernel function, which is preferred mainly for input features with non-linear relationships as it transforms them into infinite-dimensional space. Equation \eqref{eqn:rbf} gives the expression for the Gaussian RBF function:
\begin{equation}
\label{eqn:rbf}
K(x, x_i) = \exp \big(\frac{-(x - x_i) ^ 2}{2\sigma^2} \big)
\end{equation}

here $\sigma$ is the width of the core, also called the kernel parameter. The error penalty parameter $\gamma$ and the width of the core $\sigma$ are important parameters of LSSVM, possessing a great impact on the overall performance of the algorithm. The estimation of a good prediction model using a finite set of training data requires maintaining a trade-off between achieving a small bias and a small variance. In LSSVM, this trade-off is maintained using the error penalty ($\gamma$). Choosing a very large value of $\gamma$ causes the model to overfit, learning both the patterns and noise in the data, resulting in a model with poor generalization capabilities. Very small values of $\gamma$ cause the model to underfit, failing to learn any patterns from the training set itself \cite{lendasse2005ls}. Moreover, the choice of kernel parameter $\sigma$ provides important implications for the performance of the LSSVM algorithm. The kernel parameter defines the structure of the high dimensional feature space, which is unevenly distributed among the patterns in the training data. Since these feature spaces tend to overlap, a very low value of $\sigma$ causes the model to overfit as it is unable to separate the patterns. When a very high value of $\sigma$ is chosen, the model underfits since all patterns appear to be similar \cite{chang2005scaling}. Therefore, to obtain optimum values for these parameters for LSSVM in WSF, we implemented the EBQPSO algorithm.

\subsection{QPSO with elitist breeding \label{secebqpso}}
QPSO is a novel particle swarm optimization (PSO) algorithm that works by emulating the quantum mechanics of particles. It updates the particle positions using the delta potential model in the vicinity of the previous particles \cite{sun2011quantum} and utilizes the mean best position to improve its global search capabilities \cite{sun2004global}. The particles of QPSO move in spinless quantum space, and their positions are estimated using a probability density function \cite{zhou2008qpso}. The possible space is then probed in search of the most optimal point. Given a set of $M$ particles, with each defined in an $n$ dimensional space, the position of particle $i$ during iteration $t$ is given as $X_i(t) = (X_{i, 1}(t), \dots, X_{i, n}(t))$ with its evolution governed using the following three equations:
\begin{equation}
\label{eqn:mbest_1}
m_{best}(t) = \frac{1}{M}\sum_{i=1}^M pbest_{i}
\end{equation}
\begin{equation}
\label{eqn:pci}
P_{c_{i, j}}(t) = \phi_{i, j}(t)*pbest_{i, j}(t) + (1 - \phi_{i, j}(t)) * gbest_j(t)
\end{equation}
\begin{equation}
\label{eqn:xit}
X_{i, j}(t+1) = P_{c_{i, j}}(t) \pm a*|m_j - X_{i, j}(t)|\ln(\frac{1}{u})
\end{equation}
$pbest_i(t)$ is the personal best position, which is the best position of particle $i$ from previous iterations, which is the particle with the best fitness value. $m_{best}$ is the mean of the personal best positions of all particles given as:
\begin{equation}
\label{eqn:mbest_2}
m_{best} (t) = \big( \frac{1}{M} \sum_{i=1}^M pbest_{i, 1} (t) \dots \frac{1}{M} \sum_{i=1}^M pbest_{i, n} (t) \big)
\end{equation}
$gbest(t)$ is the global best position, i.e. the best position of all particles. $P_{c_{i}}$ is a random position between $pbest_i(t)$ and $gbest(t)$, $\phi$ and $u$ are any random numbers between 0 and 1 inclusive; $\alpha$ is the contraction expansion (CE) coefficient, which is tuned to alter the speed of convergence of QPSO. For the $t^{th}$ iteration, it is given as:
\begin{equation}
\label{ce}
\alpha = 0.5 + 0.5 * (T - t)/T
\end{equation}
where $T$ is the maximum number of iterations. 

\begin{algorithm}[!ht]
\caption{Procedure for transposon operator}
\begin{algorithmic}[1]
\label{alg:transposonop}
\Procedure{TransposonOperator}{$epool$}
        \LineComment{$epool$ is pool of particles}
        \State Define population size $M$, number of transposon $L$, and jumping rate $jrate$
        \State Generate $epool\_norm$ from $epool$ based on equation \eqref{norm}
        \For{$i = 1$ to $M+1$}
                \If{random(0, 1) $<$ $jrate$}
                        \State $C_1 = i$
                        \State $C_2 = ceil(rand(0,1) \times (M+1))$
                
                        \If{$C_1 == C_2$}
                                \If{random(0, 1) $>$ 0.5}
                                        \State Apply cut and paste operation in $epool\_norm[C1]$
                                \Else
                                        \State Apply copy and paste operation in $epool\_norm[C1]$
                                \EndIf
                        \Else
                                \If{random(0, 1) $>$ 0.5}
                                        \State Apply cut and paste operation in $epool\_norm[C1]$                                                       \State and $epool\_norm[C2]$
                                \Else
                                        \State Apply copy and paste operation in $epool\_norm[C1]$
                                        \State and $epool\_norm[C2]$
                                \EndIf
                        \EndIf
                \EndIf
        \EndFor
\EndProcedure
\end{algorithmic}
\end{algorithm}

The PSO algorithm first initializes random solutions (or particles) and iteratively updates these solutions to find an optimal solution \cite{zhou2008qpso}. This makes it different from other evolutionary algorithms, whose solutions depend on evolutionary operations such as crossover and mutation. PSO relies on the previous solutions of its particles, which include the position and velocity vectors, while searching the problem space \cite{sun2012quantum}. It is also influenced by parameters that are easily adjustable. These characteristics make the algorithm computationally cheap, easy to implement, and applicable for several real-world optimization problems \cite{zhang2015comprehensive}. However, PSO's performance is reduced when applied to complex multimodal optimization tasks, getting easily stuck in local optimum values \cite{tian2018mpso}.

QPSO, on the other hand, excludes the velocity vector and contains fewer adjustable parameters than PSO. This results in a simplified optimization scheme with more suitability for real-world problems. The parameters of QPSO have a direct impact on the performance of the algorithm. Equation \ref{eqn:mbest_1} computes the value of $m_{best}$, which is the average optimal position of the whole population, which is particularly important for the performance of the algorithm. $m_best$ evaluates the mean of the personal best positions of all particles. This takes into account the influence of each particle in the population equally in obtaining the final solution. However, considering all particles equally can negatively impact the overall solution as some particles have more influence (or varying elitism) than others in finding the optimal solution.

The personal best and the global best are merely used for updating particle positions, despite their elitist role in the algorithm. However, further manipulation of these elitist particles may produce better solutions. Therefore, in this paper, a QPSO algorithm with elitist breeding (EBQPSO) is developed to enhance the performance of QPSO and utilize it to search for optimal parameters of LSSVM. By forming a new swarm of elitists through breeding, EBQPSO attempts to share new potential solutions to the regular QPSO algorithm. This is achieved by first constructing an elitist pool, $epool$, consisting of the latest personal best and global best particles. Then, a breeding scheme, known as the transposon operator, is implemented to generate more diversified and optimal solutions.

Elitist breeding mimics the mutation of biological DNA elements known as transposons. Transposons are a sequence of genes, with each location allocated randomly in chromosomes. Transposon operators are operations that move genes from one position to another in a single or between two chromosomes. Transposon operators are classified into two categories: cut and paste operators and copy and paste \cite{yang2015improved}. Choosing which operation to apply is performed randomly. The size of the transposon is greater than or equal to one and is chosen using a parameter called jumping percentage. The transopon operation is not applied in each iteration but is activated only in certain conditions that are activated using a parameter called jumping rate. The operation is performed between the current and some randomly chosen chromosome when the jumping rate is greater than or equal to some randomly selected number between one and zero. If this results in the position of the current chromosome being equal to the position of the randomly selected chromosome, it signifies the operation is to be applied to the same chromosome; otherwise, it is done on different chromosomes. Moreover, the choice of whether to apply cut-and-paste or copy-and-paste is determined randomly. The procedure for the transposon operator is given in algorithm 1.

\begin{algorithm}[!ht]
\label{alg:ebqpso}
\caption{Procedure EBQPSO}
\begin{algorithmic}[1]
\State Define search space, population and fitness function $f$
\State Set population size $M$, dimension $n$, max iteration $T$
\State Generate random particles $\textbf{X}$
\State Initialize $pbest[i] = X[i] \quad 1 \leq i \leq M$
\For{$t = 1$ to $T$}
        \State $gbest = \argmin(f(pbest))$
        \State Compute $m_{best}$ based on equation \eqref{eqn:mbest_2}.
        \If{elitist criterion met}
                \State $i = 1$
                \While{$i \leq M$}
                        \State $epool[i] = pbest[i]$
                        \State $i = i+1$
                \EndWhile
                \State $epool[M+1] = gbest$
                \State $epool\_eb=TransposonOperator(epool)$
                \For{i=1 to M}
                        \If{$f(epool\_eb[i]) < f(pbest[i])$}
                                \State $pbest[i] = epool\_eb[i]$
                        \EndIf
                \EndFor
                \State $gbest = argmin(f(pbest)$
        \EndIf
        \For{i=1 to M}
                \State Compute $P_{c_i}$ based on equation \eqref{eqn:pci}.
                \State Update $X[i]$ based on equation \eqref{eqn:xit}.
                \If{($f(X[i])<f(pbest[i])$}
                        \State $pbest[i] =X[i]$
                \EndIf
        \EndFor
\EndFor
\end{algorithmic}
\end{algorithm}

Each particle in QPSO is regarded as a chromosome, and each gene in the chromosome is regarded as the particle positional value. Therefore, the number of chromosomes is the same as the number of particles in the swarm. The dimensional values of each particle can have different search spaces. Thus, they are normalized as follows: \begin{equation} \label{norm} \textbf{x}_{norm} = \frac{\textbf{x} - max(\textbf{x})}{max(\textbf{x}) - min(\textbf{x})} \end{equation} where $min(\textbf{x})$ and $max(\textbf{x})$ represent the lower and upper bounds of $\textbf{x}$, respectively. After the application of the transposon operator, the positional values of the output vector are transformed back to the original positional value within the search space using the following equation: \begin{equation} \label{denorm} \textbf{x} = \textbf{x}_{norm} \times (max(\textbf{x}) - min(\textbf{x})) + min(\textbf{x}) \end{equation}

When the criteria for transposon operation is met, the algorithm produces a new sub-swarm, $epool_eb$. The $pbest$ is modified if the fitness value in the corresponding $epool$ is better. To control how many times elitist breeding is carried out, a predefined hyper-parameter called lambda ($\lambda$) is used, by which breeding is performed in every $\lambda$ iteration. Algorithm 2 illustrates the overall EBQPSO algorithm.

Table \ref{tab:complexity} illustrates the time and memory complexity of PSO, QPSO, and EBQPSO in terms of number of generations ($T$), population size ($N$), dimension ($D$), and lambda ($\lambda$). The memory complexity of each algorithm is $\Theta(N*D)$. However, while the time complexity of PSO and QPSO is $\Theta(T*N*D)$, EBQPSO's running time is $\Theta(\frac{2 + \lambda}{\lambda} * T*N*D)$, slightly larger by a factor of $\frac{2+\lambda}{\lambda}$ approximately a constant number. Despite the difference in running time, EBQPSO is in general more robust for optimizing machine learning parameters, mainly due to its faster convergence. This enhanced convergence capability is a result of the utilization of elitist particles using the transposon operator. By exploiting the elite particles early during the iterative process, EBQPSO produces optimum values faster. This characteristic is suitable for searching for the for the optimum hyperparameter values of machine learning models, especially when the dataset size is large, as the algorithm converges using fewer generations than would be required for PSO or QPSO.

\begin{table}[!ht]
\centering
\caption{ Average time and memory complexity comparison}
\begin{tabular}{|l|c|c|}
\hline
\textbf{Algorithm} & \textbf{Time complexity} & \textbf{Memory complexity} \\\hline
PSO & $\Theta(T*N*D)$ & $\Theta(N*D)$  \\\hline
QPSO & $\Theta(T*N*D)$ & $\Theta(N*D)$ \\\hline
EB-QPSO & $\Theta(\frac{2 + \lambda}{\lambda} * T*N*D)$ & $\Theta(N*D)$ \\\hline
\end{tabular}
\label{tab:complexity}
\end{table}

\subsection{Long short-term memory (LSTM)}
LSTM is an upgrade over RNN to address the problems of vanishing or exploding gradients. A typical LSTM unit consists of a cell, an input gate, an output gate, and a forget gate. The function of the cell is to memorize values across arbitrary time periods, while the three gates control how much information should flow into and out of the cell. The forget gate selects which information from the prior state to pass by assigning zero or one value \cite{karim2019insights}. In every time step, the flow of information is computed as follows:

\begin{equation}
\begin{split}
    \mathbf{f_t} = \mathbf{\mbox{sigmoid}(W_f \cdot [h_{t-1}, x_t] + b_f)}     \\
    \mathbf{i_t} = \mathbf{\mbox{sigmoid}(W_i \cdot [h_{t-1}, \, x_t] + b_i)} \\    
    \mathbf{\tilde{C_t}} = \mathbf{\mbox{tanh}(W_C \cdot [h_{t-1}, \, x_t] + b_C)} \\
    \mathbf{C_t} = \mathbf{f_t \odot C_{t-1} + i_t \odot \tilde{C_t}} \\
    \mathbf{o_t} = \mathbf{\mbox{sigmoid}(W_o \cdot [h_{t-1}, \, x_t] + b_o)} \\
    \mathbf{h_t} = \mathbf{o_t \odot tanh(C_t)}
\end{split}    
\end{equation}

where $f_t$, $i_t$, $\tilde{C_t}$, and $o_t$ are the activation vectors of the input, forget, output and cell state gates respectively, $W_f, W_i, W_c$, and $W_o$ are weight matrices, and $b_f, b_i, b_c$, and $b_o$ denote the bias vectors. $C_t$ is the memory vector used to regulate the outputs and state updates, and $h_t$ is the hidden vector at time step $t$. $\odot$ represents element-wise multiplication, $\mbox{sigmoind}(\cdot)$, also denoted by $\sigma(\cdot)$, is the logistic sigmoid function which is utilized as the gate activation function:
\begin{equation}
    \sigma(x) = \frac{1}{1 + e^{-x}}
\end{equation}

and $tanh(\cdot)$ is a hyperbolic tangent function used as an activation function:

\begin{equation}
    \mbox{tanh(x)} = \frac{e^x - e^{-x}}{e^x + e^{-x}}
\end{equation}

\begin{figure}
    \centering
    \includegraphics[scale=.5]{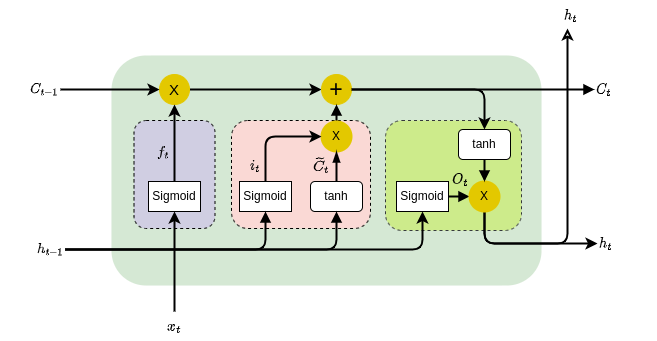}
    \caption{LSTM architecture used for error sequence modeling.}
    \label{fig:lstm_arch}
\end{figure}
\subsection{Dataset}
Figure \ref{fig:lstm_arch} a typical LSTM architecture. In this work, LSTM is used to model the deviation between the sum of the intrinsic modes and the original wind series. The gating mechanisms and ability to handle vanishing and exploding gradients make LSTM well-suited for capturing non-linear and non-stationery characteristics of the error sequence. 

\subsubsection{Dataset acquisition and description}
The dataset used for this study is collected from Ashegoda Wind Farm, located south of Mekelle City, Ethiopia, at 13$^\circ$ 25' 16.8" N, 39$^\circ$ 34' 20.7" E. Ashegoda Wind Farm is a 120 MW onshore wind farm operating 84 wind turbines, making it the largest wind farm in sub-Saharan Africa. The wind speed dataset, along with other relevant information, is obtained from the supervised control and data acquisition (SCADA) system. In our study, we consider two The first data was collected from April 1 to 30, 2015, constituting a total of 1400 sample points, as shown in Table \ref{tab:dataset_summary}. The first data was collected from April 1 to 30, 2015, constituting a total of 1400 sample points. The second set of data was collected from May 1 to May 31, 2015, making up a total of 1448 samples. Each point in both datasets was sampled every 20 minutes.

\begin{table}[!ht]
    \centering
    \caption{Dataset description and summary}
    \begin{tabular}{llllll}
        \hline
        Dataset &  Number of points & Min. & Max. & Mean & Std \\\hline
        Dataset-1 (April Dataset) & 1440 & 1.02 & 13.67 & 9.04 & 1.83 \\
        Dataset-2 (May Dataset) & 1448 & 0.37 & 12.14 & 5.81 & 2.09 \\\hline
    \end{tabular}
    \label{tab:dataset_summary}
\end{table}
\begin{figure}[!ht]
\centering
\includegraphics[scale=.15]{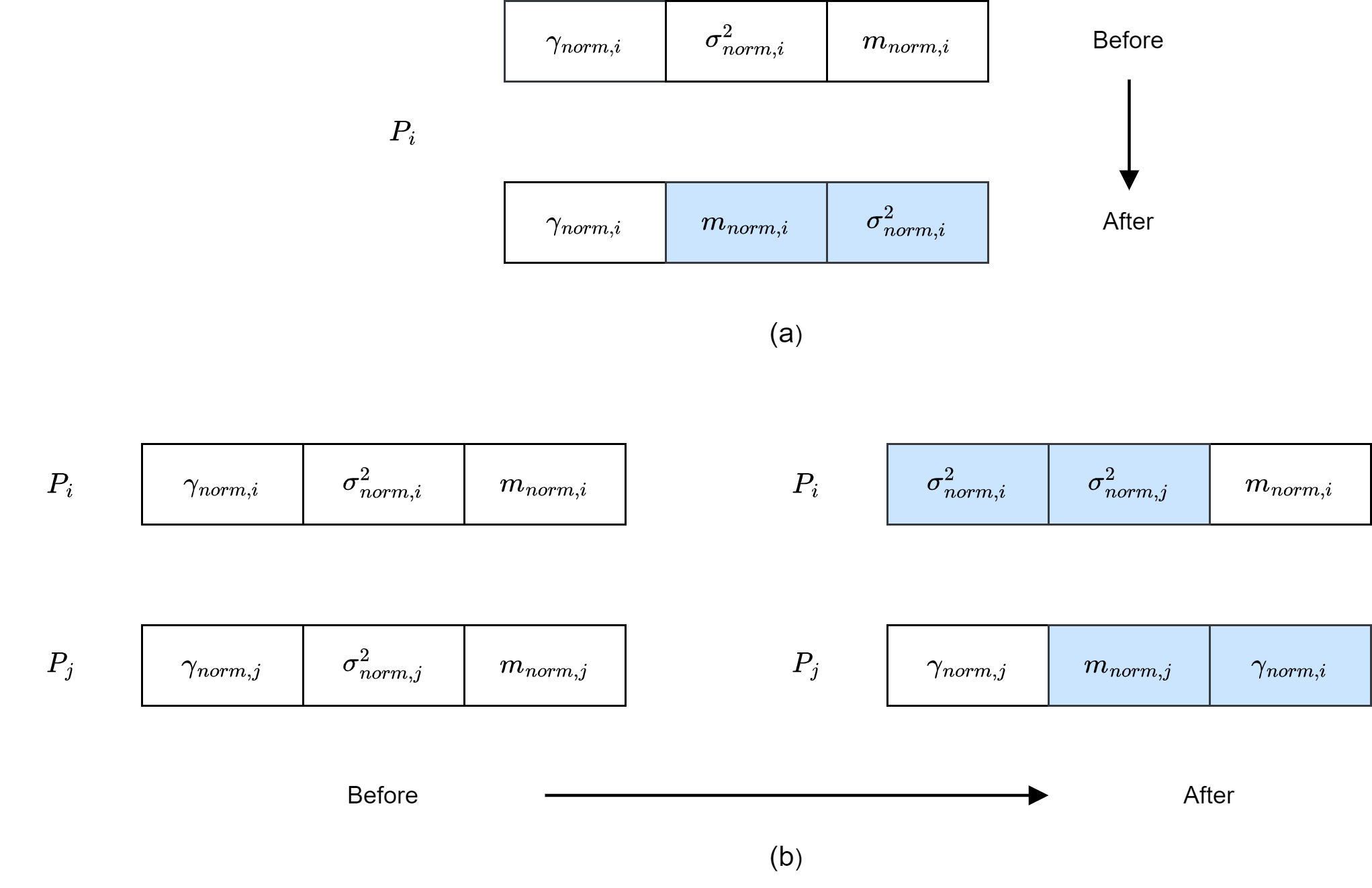}
\caption{\label{fig:cutpaste} Applying the cut-and-paste transposon operator to LSSVM parameters and window size: (a) Same particle; (b) Different particles}
\end{figure}
\begin{figure}[!ht]
\centering
\includegraphics[scale=.15]{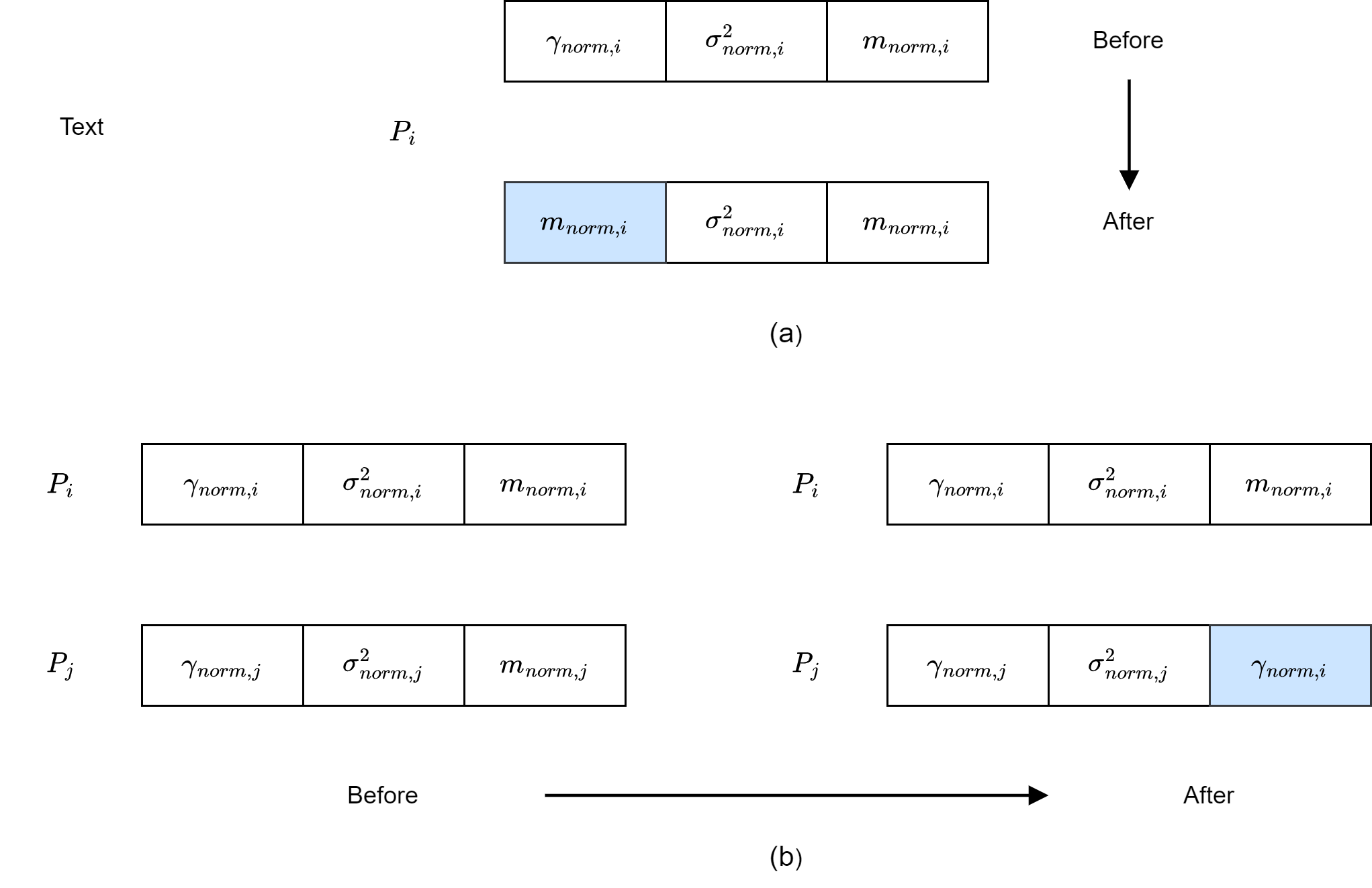}
\caption{\label{fig:copypaste} Applying the copy-and-paste transposon operator to LSSVM parameters and window size: (a) Same particle; (b) Different particles}
\end{figure}
\subsection{Data cleaning and feature selection}
The SCADA system can sometimes generate missing values and outliers due to faulty sensors and erroneous measurements. These inaccurate measurements have a negative impact on the performance of the prediction model. Thus, handling them vigorously is important. The May and April datasets used in this work contain very few missing points in comparison to the size of the data. Therefore, we replaced the missing data and outliers with the mean value of the training wind data. Building a forecasting model for time series applications includes transforming the original wind time series into a supervised learning task. Taking into account the wind data, a time series represented as a univariate sample with a sampling frequency of 20 minutes given as $\{x_1, x_2, \dots\}$, a feature space of input outputs in the form of $D := ((x_1, y_1), \dots , (x_N, \dots, y_N)$ is constructed. We have $x \in \mathbb{R}^m$ and $y \in \mathbb{R}$ where $m$ is the number of features to be selected. Wind speed can be effectively forecast using its own historical data with LSSVM. The SVMD algorithm is first implemented to decompose the initial wind speed series into its constituent intrinsic modes. Each mode is normalized, and the EBQPSO algorithm is used to generate the optimum window size $m_{opt}$ with which the input features are created. This means the window size is another hyperparameter whose optimum value is searched by the EBQSPO algorithm. Therefore, the goal of the EBQPSO algorithm is to search for the best values of the kernel parameter, the regularization parameter, and the window size for each mode of the decomposed series.

\subsection{Modeling procedure of the proposed method}
This study proposes the SVMD-EBQSPO-LSSVM-LSTM method for short-term wind speed forecasting. Each modal component generated by SVMD is modeled by the EBQSO-optimized LSSVM algorithm. And for the error sequences, an LSTM method is used. To implement QPSO with elitist breeding (EB-QPSO) for the search of optimum values of LSSVM hyper-parameters and the window size, first the parameters have to be encoded as chromosomes. The regularization parameter ($\gamma$), the kernel parameter ($\sigma ^2$), and the window size should contain three positional values, signifying each chromosome as a particle with three genes and restricting the number of transposons and the size of each transnsposon to one. The cut-and-paste and copy-and-paste operations can be summarized as in Figures \ref{fig:cutpaste} and \ref{fig:copypaste}.

\begin{figure}[!ht]
\centering
\includegraphics[scale=.11]{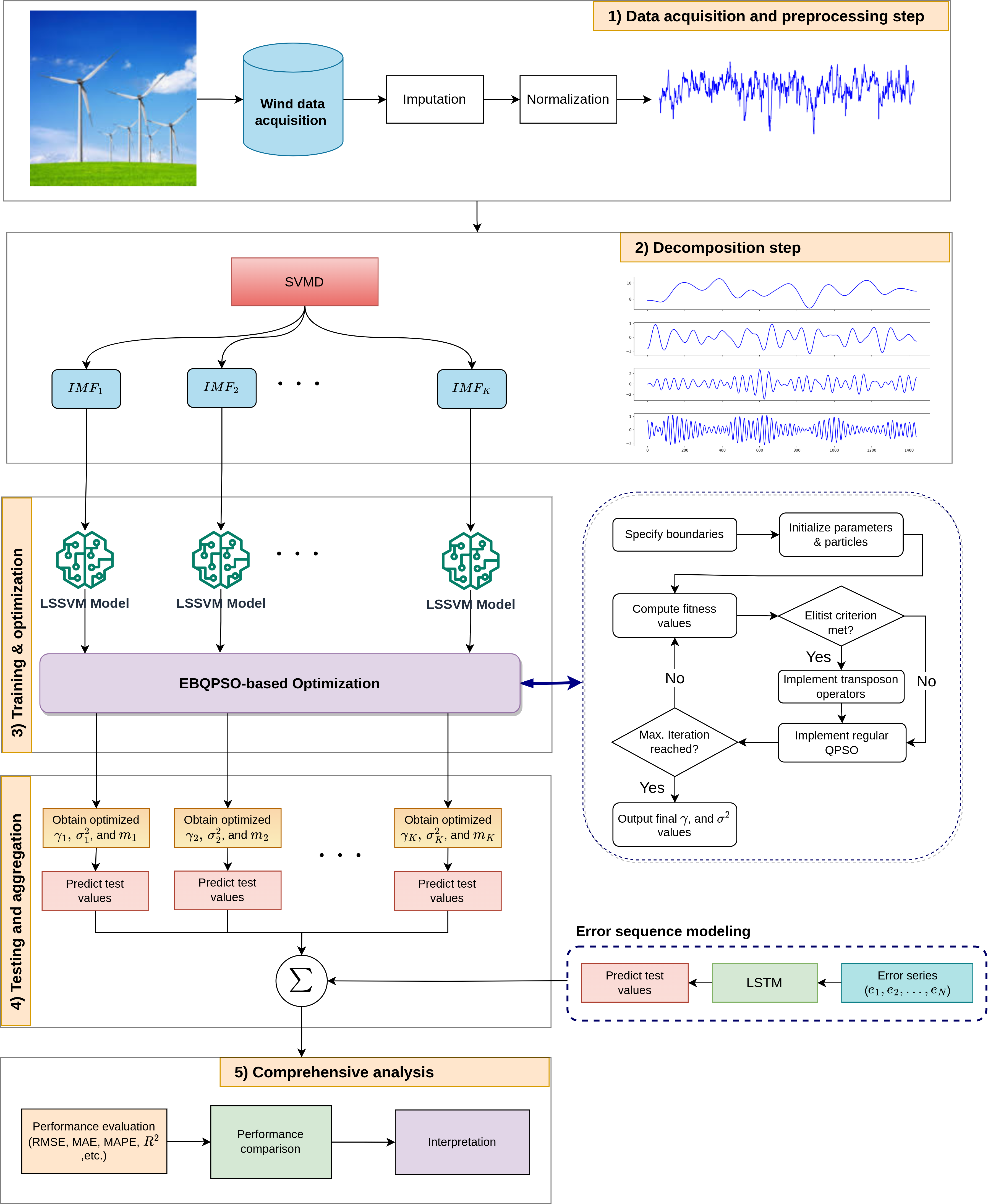}
\caption{\label{fig:methodology} Illustration of the procedures followed for the proposed method.}
\end{figure}

Further, the selection of a good fitness function is a key component of successfully implementing evolutionary algorithms. In this paper, the inverse of the mean square error (MSE), given in equation \ref{eqn:fitness}, is selected as the fitness function of the EB-QPSO algorithm. This is mainly due to the computational simplicity of MSE compared to other metrics such as the root mean squared error (RMSE).
\begin{equation}
\label{eqn:fitness}
f_{mse}(\mathbf{y}, \mathbf{\hat{y}}; \gamma, \sigma^2, m) = \frac{1}{MSE(\mathbf{y}, \mathbf{\hat{y}}; \gamma, \sigma^2, m)} = \frac{1}{N}\sum_{i=1}^N |\mathbf{y}(i) - \mathbf{\hat{y}}(i)|^2
\end{equation}

Therefore, the SVMD-EBQPSO-LSSVM-LSTM model first optimizes the LSSVM parameters and the window size using each mode as training and validation sets by maximizing the fitness function (i.e., the inverse of mean squared error). Then the LSTM network models the error sequence that makes up for the differences between the aggregate of the decomposed modes and the original wind series. The final prediction results are computed by summing the prediction values of the intrinsic modes and the error values. Essentially, the basic steps involved in training, optimizing, and testing our proposed model are given as follows:

\begin{itemize}
\item[] \textbf{Step 1:} Access the wind data from the SCADA system. Check for missing values and replace them with the mean of the dataset.
\item[] \textbf{Step 2:} Use the proposed SVMD decomposition scheme to break down the wind data into intrinsic modes. This work uses two datasets, the April and May datasets, to experiment with the proposed model. The SVMD method is applied to both datasets, and the decomposed modes are shown in Figures \ref{fig:vmds1} and \ref{fig:vmds2}.
\item[] \textbf{Step 3:} For each decomposed mode, split the series for training (70\%), validation (15\%), and testing (15\%), and implement the EBQPSO-LSSVM algorithm as follows:.
\begin{itemize}
\item[] \textbf{Step 3.1:} Define the maximum number of iterations $T$ and the population size $M$ of three dimensions as $X_i = (\gamma_i, \sigma_i^2, m_i)$. Choose the value of lambda.
\item[] \textbf{Step 3.2:} At every iteration $i$, use the window size $m_i$ to create input features and target values for the training and validation sets. That is, if the original series is $X = (x_1, x_2, \dots, x_N)$, the newly created feature input and target output based on $m$ historical points are given as:
\[ 
\mathbf{X}  = \begin{bmatrix}
				x_1 & x_2 & \dots & x_m \\
				x_2 & x_3 & \dots & x_{m+1} \\
				\vdots & \vdots & \ddots & \vdots \\
				x_{N-m} & x_{N-m+1} & \dots & x_{N-1}
			 \end{bmatrix} 
\qquad
\mathbf{y} = \begin{bmatrix}
x_{m+1} \\
x_{m+2} \\
\vdots \\
x_{N}
\end{bmatrix}
\]
\item[] \textbf{Step 3.3:} Use the training set to train LSSVM with regularization parameter $\gamma_i$ and kernel parameter $\sigma^2_i$, and the validation set to compute the fitness value of each particle according to equation \ref{eqn:fitness}. 
\item[] \textbf{Step 3.4:} Using the computed fitness value, update $pbest$ for each particle and $gbest$ for the entire. Evaluate $m_{best}$ using equation \ref{eqn:mbest_2}. 
\item[] \textbf{Step 3.5:} When lambda becomes divisible by the current iteration value, construct $epool$ by merging $pbest$ and $gbest$ into one pool. Apply the transposon operation to $epool$. Update $pbest$ and $gbest$ by evaluating the fitness of the newly bred values in the pool. Otherwise, go to step 3.6. 
\item[] \textbf{Step 3.6:} Update the particles of the swarm using equations \ref{eqn:pci} and \ref{eqn:xit}. 
\item[] \textbf{Step 3.7:} Check if the condition for optimization is met. If the maximum number of iterations is reached, end the optimization process and generate the optimal particle positions: $(\gamma_i, \sigma_i^2, m_i)$. Otherwise, return to step 3.2. 
\item[] \textbf{Step 3.8:} Output the final optimized parameters $\gamma_{opt}$, $\sigma_{opt}^2$, and $m_{opt}$. Use $m_{opt}$ to create input features and target values for the training and testing samples. And retrain the LSSVM model with parameters $\gamma_{opt}$ and $\sigma_{opt}^2$ using the training sample. Test the trained model on the testing samples and output the predicted values. \end{itemize} 
\item[] \textbf{Step 4:} Compute the error sequences using the decomposed SVMD modes and the original wind series. Given the original wind speed series $\mathbf{X}$ and $L$ intrinsic modes $\{ IMF_1, IMF_2, \dots, IMF_L \}$ all of size $N$, the error sequence $\mathbf{E}$ is given as:
\begin{equation}
\label{eqn:err_seq}
\mathbf{E = X - \sum_{i=1}^L IMF_i}
\end{equation}
\item[] \textbf{Step 5:} Split the error sequence into training, validation, and testing samples and convert the samples into input-output features using a window size of 5. Use an LSTM network to train and validate the model. Use the testing sample to test the LSTM model and output the predicted values of the error sequence. The parameters used for modeling the LSTM are shown in table \ref{tab:comp_meth_params}. 
\item[] \textbf{Step 6:} The final predicted values of the wind speed series are obtained by taking the aggregate of the predicted values of each decomposed mode and the error sequence. 
\end{itemize} 
One challenge while taking the aggregate is that the window sizes used for each decomposed value and the error sequence can be different, resulting in a length mismatch. The solution to this problem is to first compute the difference between the maximum window size and the window size selected for a given series. Assuming that difference is denoted as $d_i$, the first $d_i$ values of that series are divided, resulting in a length match for all the series to be added. The overall illustration of the modeling process of the proposed system is depicted in Figure \ref{fig:methodology}.
Many standards for evaluating the performance of prediction models are known. 

In this work, we use the mean absolute error (MAE), the root mean square error (RMSE), and the mean absolute percentage error (MAPE) given as follows:
\begin{equation}
\label{eqn:mae}
MAE = \frac{1}{N}\sum_{i=1}^N |y_i - \hat{y}_i|
\end{equation}
\begin{equation}
\label{eqn:rmse}
RMSE = \sqrt{\frac{1}{N}\sum_{i=1}^N |y_i - \hat{y}_i|^2} 
\end{equation}
\begin{equation}
\label{eqn:mape}
MAPE = \frac{1}{N}\sum_{i=1}^N \Big|\frac{y_i - \hat{y}_i}{y_i}\Big| \times 100
\end{equation}
\begin{equation}
    \label{eqn:r2}
    R^2 = 1- \frac{\sum\limits_{i=1}^N (y_i - \hat{y_i})^2}{\sum\limits_{i=1}^N (y_i - \bar{y})^2}
\end{equation}
\begin{equation}
	\label{eqn:cc}
	CC = \frac{\sum\limits_{i=1}^{n} (y_i - \overline{y}) (\hat{y_i} - \overline{\hat{y}})} {\sqrt{\sum\limits_{i=1}^{n} (y_i - \overline{y})^2 \sum\limits_{i=1}^{n} (\hat{y_i} - \overline{\hat{y}})^2}}
\end{equation}
where $N$ is the size of the test set, $y_i$ is the $i^{th}$ sample of the test set, $\overline{y}$ is the mean of the test set, $\hat{y_i}$ is the $i^{th}$ instance of the predicted series, $\overline{\hat{y}}$ is the mean of the predicted series.

\section{Results and discussion}
\label{sec:results}
\subsection{Performance evaluation of EBQPSO algorithm using benchmark functions}
To compare the performance of the EBQPSO optimization algorithm with PSO and QPSO, four well-known benchmark functions are considered. The functions are Sphere ($F_1$), Ackley ($F_2$), Griewank ($F_3$), and McCormick ($F_4$), and their mathematical expressions are given as follows:

\begin{equation}
\label{eqn:sphere}
F_1(x) = \sum_{i=1}^d x_i ^ 2
\end{equation}
\begin{equation}
	\label{eqn:ackley}
	F_2(x) = -20 \cdot \exp \left( -0.2 \cdot \sqrt{\frac{1}{d} \sum\limits_{i=1}^{d} x_i^2} \right) 
- \exp \left( \frac{1}{d} \sum\limits_{i=1}^{d} \cos(2\pi \cdot x_i) \right) + 20 + e
\end{equation}
\begin{equation}
	\label{eqn:griewank}
	F_3(x) = \sum_{i=1}^{d} \frac{x_i^2}{4000} - \prod_{i=1}^{d} \cos \left( \frac{x_i}{\sqrt{i}} \right) + 1
\end{equation}
\begin{equation}
	\label{eqn:mccormick}
	F_4(x) = \sin(x_1 + x_2) + (x_1 - x_2)^2 - 1.5 * x_1 + 2.5*x_2 + 1
\end{equation}
Each experiment was conducted five times to produce statistically confident results about its performance. The number of iterations taken was 100, and the number of populations was 25. The dimensions $d$ of functions $F_1$, $F_2$, and $F_3$ were set to be 20, while that of $F_4$ is 2. Table \ref{tab:optalgo} summarizes the final results of each algorithm in terms of mean and standard deviation as compared to the global minimum values of the benchmark functions. As can be observed from the table, EBQPSO produced the closest minimum values to the global minimum values of functions $F_1, \, F_2$, and $F_4$, while generating the same minimum value as QPSO for $F_4$. We believe this performance difference can be enhanced if more trials are used.

\begin{figure}[!ht]
    \centering
    \subfloat[]{{\includegraphics[scale=.35]{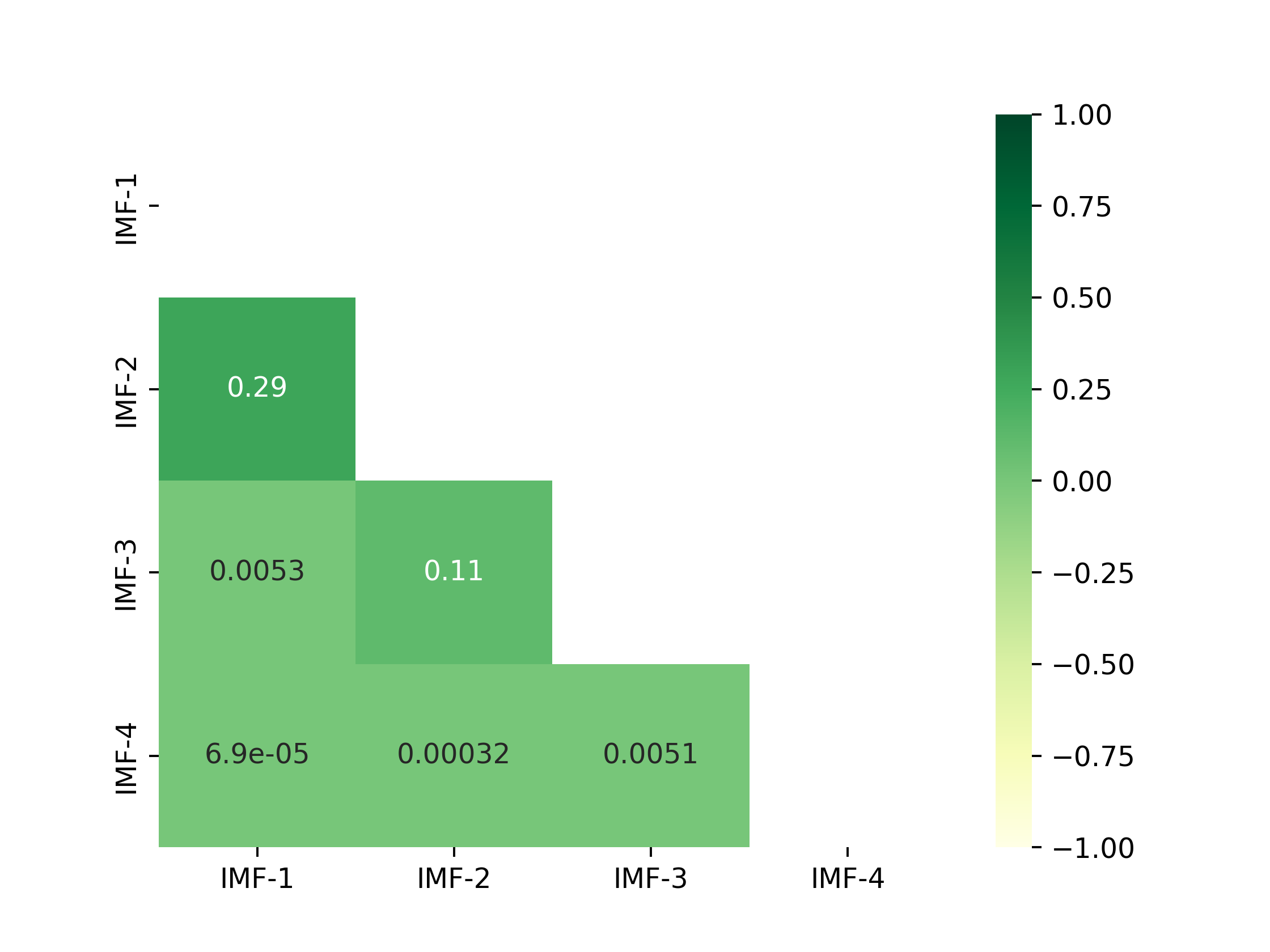} }\label{fig:imf1}}%
    \qquad
    \subfloat[]{{\includegraphics[scale=.35]{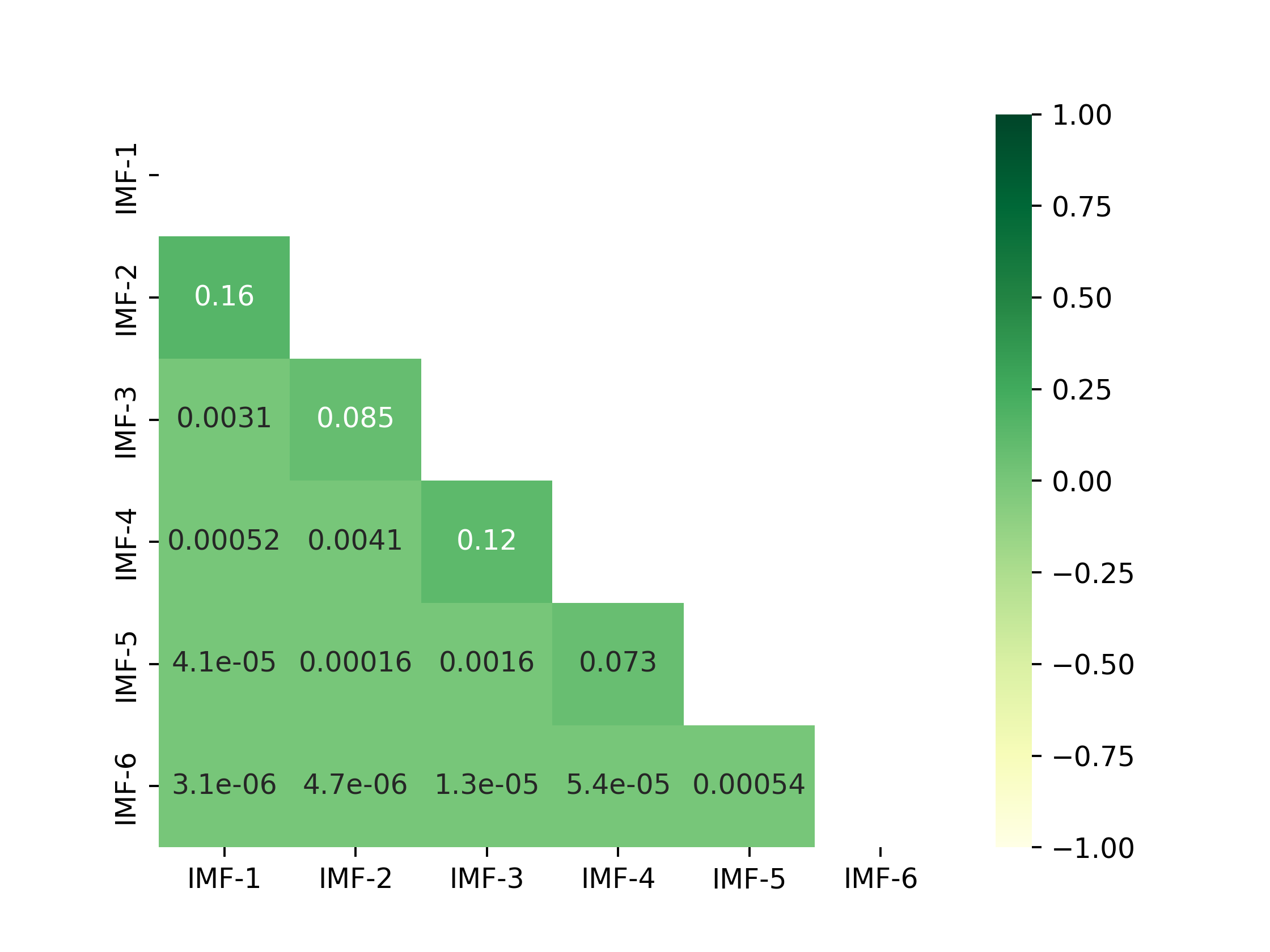} }\label{fig:imf2}}%
    \caption{\centering Diagonal Correlation table of SVMD decomposed modes: a) April dataset b) May dataset}%
    \label{fig:imf}%
\end{figure}

\begin{table}
\centering
\caption{Performance comparison of optimization methods using four benchmark functions and five trials}
\resizebox{\textwidth}{!}{
\label{tab:optalgo}
\begin{tabular}{ccccc}
\hline
\multirow{2}{*}{\textbf{Function}} & \textbf{Global} & \multicolumn{3}{c}{\textbf{Output min. value}} \\\cline{3-5}
 & \textbf{minimum} & \textbf{EBQPSO} &\textbf{QPSO} & \textbf{PSO} \\\hline
 $F_1 \, (d = 20)$ & 0.0 & $\mathbf{1.3 \times 10^{-47} \pm 1.2 \times 10^{-47}}$ & $7.7\times 10^{-47} \pm 1.4 \times 10^{-46}$ & 
 $1.2\times 10^{-06} \pm 1.8 \times 10{-06}$  \\\hline
 $F_2 \, (d = 20)$ & 0.0 & $\mathbf{1.2 \times 10^{-15} \pm 1.4 \times 10^{-15}}$ & $3.3 \times 10^{-15} \pm 1.4 \times 10^{-15}$ & $4.2 \pm 3.6$ \\\hline
 $F_3 \, (d = 20)$ & 0.0 & $0.0 \pm 0.0$ & $0.0 \pm 0.0$ & $0.18 \pm 0.22$ \\\hline
 $F_4 \, (d = 2)$ & -1.9133 & $\mathbf{-1.9132 \pm 6.8 \times 10^{-06}}$ & $-1.9130 \pm 0.000185$ & $-10306.9 \pm 5873.7$ \\\hline
\end{tabular}}
\end{table}

\subsection{Experimental setup}
Table \ref{tab:ebqpso_params} presents the parameter settings for the EBQPSO algorithm when implemented to optimize LSSVM parameters and the window size of each modal component. We set the maximum number of generations to 100 and the population size to 25. Since we are optimizing the three hyperparameters, the problem dimension is set to 3, jumping percentage to 1, and the number of transposons is also set to 1. The jumping rate is chosen to be 0.3, indicating the transposon operator is activated with a 0.30 probability; otherwise, the algorithm continues with regular QPSO. The $\lambda$ value is set to 5 to initiate elitist breeding every 5 generations. Selecting an appropriate search space is also crucial: we set the minimum value of both $\gamma$ and $\sigma^2$ to be 0.0001 and their maximum to be 10000. The minimum value of the window size is 1 and the maximum is 25. Also, we mapped the search spaces of $\sigma^2$ and $\gamma$ to log-scale to improve search capability and help converge to optimum values with a smaller number of iterations. These EBQPSO configurations are used to optimize the LSSVM model while being trained on all the decomposed modes of the April and May datasets. 

\begin{table}[!ht]
\centering
\caption{\label{tab:ebqpso_params} Parameter setting for EBQPSO algorithm}
\begin{tabular}{|l|c|}
\hline
\textbf{Parameter} & \textbf{Parameter value} \\\hline
Number of generations (G) & 100 \\\hline
Population size (M) & 25 \\\hline
Problem dimension (d) & 3 \\\hline
CE coefficient ($\alpha$) for EB-QPSO & $\alpha = 0.5$ \\\hline
Jumping percentage & 1 \\\hline
Jumping rate & 0.3 \\\hline
Number of transposons & 1 \\\hline
Lambda ($\lambda$) & 5 \\\hline
Minimum search point & ($10^{-4}$, $10^{-4}$, 1) \\\hline
Maximum search point & ($10^4$, $10^4$, 25) \\\hline
\end{tabular}
\end{table}
The error sequence is modeled using an LSTM network with parameter settings shown in Table \ref{tab:comp_meth_params}. The parameters are selected through trial-and-error to generate the best architecture. After computing the error series using equation \ref{eqn:err_seq} the LSTM is trained and tested on using the wind speed dataset.

The proposed approach is also compared with other competitive methods. The competitive methods considered in this study are LSSVM, SVMD-LSSVM, CNN, LSTM, CNN-LSTM, SVMD-CNN, SVMD-LSTM, and SVMD-CNN-LSTM. CNN, LSTM, and CNN-LSTM and their hybrid varieties have been extensively used in the literature for wind forecasting. Combining these methods, the proposed SVMD method is also a crucial approach to determining if the proposed method is superior to these benchmark models. The parameter setting of these prediction methods is illustrated in Table \ref{tab:comp_meth_params}. Moreover, by comparing the proposed method with EBQPSO-LSSVM, we can infer if including the SVMD algorithm adds any performance gain. 

\begin{table}[!ht]
\centering
\caption{Selected parameters for the competitive benchmark methods}\label{tab:comp_meth_params}
\resizebox{\textwidth}{!}{
\begin{tabular}{lll}
\toprule
\textbf{Forecasting method} & \textbf{Parameter} &  \textbf{Value} \\
\toprule
\multirow{11}{*}{CNN \& SVMD-CNN} &  Input layer nodes & 5 \\
 & 1D Convoltional output channel size (Layer 1) & 128 \\
 & Kernel size & 1 \\
 & Padding sie & 0 \\
 & Activation function & ReLU \\
 & Linear layer 1 nodes & 128 \\
 & Linear layer 2 nodes & 64 \\
 & Loss function & MSE \\
 & Learning rate & $10^{-5}$ \\
 & Optimizer & Adam \\
 & Epochs & 200 \\
\toprule
\multirow{7}{*}{LSTM \& SVMD-LSTM} & Input size & 5 \\
 & No. of features in hidden state & 200 \\
 & No. of layers & 1 \\
 & Learning rate & $10^{-5}$ \\
 & Loss function & MSE \\
 & Epochs & 500 \\
 & Optimizer & Adam \\
 \toprule
\multirow{14}{*}{CNN-LSTM \& SVMD-CNN-LSTM} & Input layer size & 5 \\
 & Layer 1 1D convlution  &  \\
 & $ \quad $ Output channel size & 64 \\
 & $ \quad $ Kernel size & 3 \\
 & $ \quad $ Stride & 1 \\
 & $ \quad $ Padding & 1 \\
 & $ \quad $ Activation function & ReLU \\
 & Layer 2 1D convolution  &  \\
 & $ \quad \quad $ Output channel, kernel size, stride, padding & (64, 3, 1, 1) \\
 & $ \quad \quad $ Activation function & ReLU \\
 & Pooling layer (Max. pool) &  \\
 & $ \quad \quad $ Kernel size, strid & (1, 1) \\
 & LSTM input size & 128 \\
 & $ \quad \quad $ No. of features in hidden state & 200 \\
 & $ \quad \quad $ No. of layers & 1 \\
 & Learning rate & $10^{-5}$ \\
 & Loss function & MSE \\
 & Epochs & 200 \\
 & Optimizer & Adam \\
 \toprule
\end{tabular}}
\end{table}

\subsection{Correlation analysis of decomposed modes of wind speed data}
The relationships among the intrinsic mode functions and the original wind series can be evaluated using correlation values. Correlation values can help us understand if the SVMD algorithm decomposes the wind speed into modes that are independent and if the center of frequency of each decomposed signal is adequately separated. Figures \ref{fig:imf1} and \ref{fig:imf2} depict the diagonal correlation matrix of IMFs for both the April and May datasets. We can observe from the figures that the highest correlation values are 0.29 and 0.16 and occur between IMF-1 and IMF-2 for both the April and May datasets, respectively. The rest of the correlation values are very small in magnitude. This small correlation output indicates that the SVMD algorithm resulted in IMFs that are independent and dissimilar. It also shows that the center frequencies of the IMFs are further away from one another, proving the effectiveness of the proposed SVMD method in decomposing the original wind speed signals.

\subsection{Performance comparison of proposed method with benchmark methods}

In this study, seven competitive models are considered to benchmark the performance of the proposed model. The methods are LSSVM-EBQPSO, CNN, SVMD-CNN, LSTM, SVMD-LSTM, CNN-LSTM, and SVMD-CNN-LSTM. Table \ref{tab:perf_comp} illustrates the performance of all methods using various metrics for both the April and May datasets. As can be seen in the table, the SVMD-EBQPSO-LSSVM-LSTM method outperformed all methods in terms of all performance metrics for both datasets.

For the April dataset, the proposed model obtained a RMSE of 0.703, a MAE of 0.512, a MAPE of 5.9\%, a R$^2$ of 0.796, and a correlation coefficient of 0.892. The lowest-performing model is the SVMD-CNN-LSTM model, while the second-highest-performing model is the SVMD-CNN model. There is a 2.42\% performance improvement by the proposed method over the second-best model in terms of RMSE, 4.10\% in terms of MAE, 3.38\% in terms of MAPE, and 1.27\% in terms of R$^2$. Moreover, a 33.85\% performance difference is obtained between the proposed and the least-performing method in terms of RMSE, 48.82\% in terms of MAE, 40.68\% in terms of MAPE, 25.35\% in terms of R$^2$, and 1.25\% in terms of correlation coefficient.

\begin{table}[!h]
\centering
\caption{Selected parameters for the competitive benchmark methods}\label{tab:perf_comp}
\begin{tabular}{c|ccccc}
\hline
 & \multicolumn{5}{c}{\textbf{April Dataset}} \\\hline
\textbf{Model} & \textbf{RMSE} &  \textbf{MAE} & \textbf{MAPE (\%)} & \textbf{$R^2$} & \textbf{CC} \\\hline
LSSVM-EBQPO &  0.785 & 0.572 & 6.6 & 0.746 & 0.865 \\\hline
Proposed & \textbf{0.703} & \textbf{0.512} & \textbf{5.9} & \textbf{0.796} & \textbf{0.892} \\\hline
CNN & 0.792 & 0.578 & 6.6 & 0.741 & 0.864 \\\hline
SVMD-CNN & 0.720 & 0.533 & 6.1 & 0.786 & 0.892 \\\hline
LSTM & 0.787 & 0.581 & 6.7 & 0.744 & 0.864 \\\hline
SVMD-LSTM & 0.784 & 0.605 & 6.7 & 0.747 & 0.89 \\\hline
CNN-LSTM & 0.809 & 0.599 & 6.8 & 0.730 & 0.862 \\\hline
SVMD-CNN-LSTM & 0.941 & 0.762 & 8.3 & 0.635 & 0.881 \\\hline
 & \multicolumn{5}{c}{\textbf{May Dataset}} \\\hline
\textbf{Model} & \textbf{RMSE} &  \textbf{MAE} & \textbf{MAPE (\%)} & \textbf{$R^2$} & \textbf{CC} \\\hline
LSSVM-EBQPSO & 1.127 & 0.877 & 16.7 & 0.683 & 0.829 \\\hline
Proposed & 0.856 & 0.661 & \textbf{13.0} & 0.817 & 0.905	\\\hline
CNN & 1.064 & 0.819 & 16.0 & 0.717 & 0.847	\\\hline
SVMD-CNN & 0.877 & 0.669 & 13.4 & 0.808 & 0.901 	\\\hline
LSTM & 1.065 & 0.816 & 15.7 & 0.717 & 0.847 \\\hline
SVMD-LSTM & 0.856 & \textbf{0.658} & 13.1 & 0.817 & \textbf{0.906} \\\hline
CNN-LSTM & 1.063 & 0.825 & 16.0 & 0.718 & 0.847 \\\hline
SVMD-CNN-LSTM & 0.899 & 0.689 & 13.8 & 0.798 & 0.896  \\\hline
\end{tabular}
\end{table}

The performance margins of the proposed method are more pronounced mainly in terms of RMSE and MAE, indicating that the SVMD-EBQPSO-LSSM-LSTM model is superior at capturing large errors and less sensitive to outliers. 

\begin{figure}[!ht]
    \centering
    \includegraphics[scale=.35]{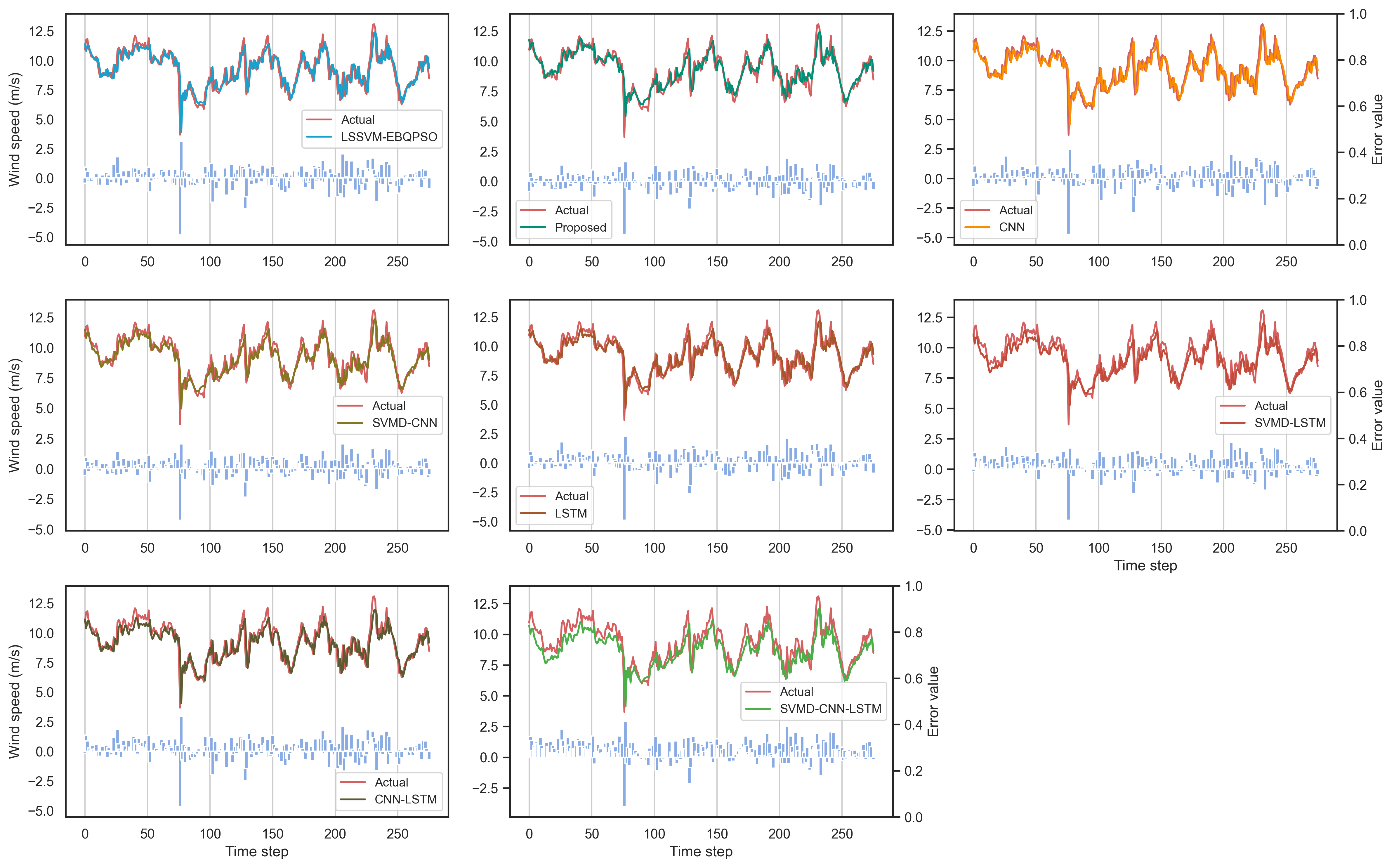}
    \caption{Actual vs Predicted results and the corresponding error bars for the April Dataset.}
    \label{fig:act_pred_1}
\end{figure}

\begin{figure}[!ht]
    \centering
    \includegraphics[scale=.35]{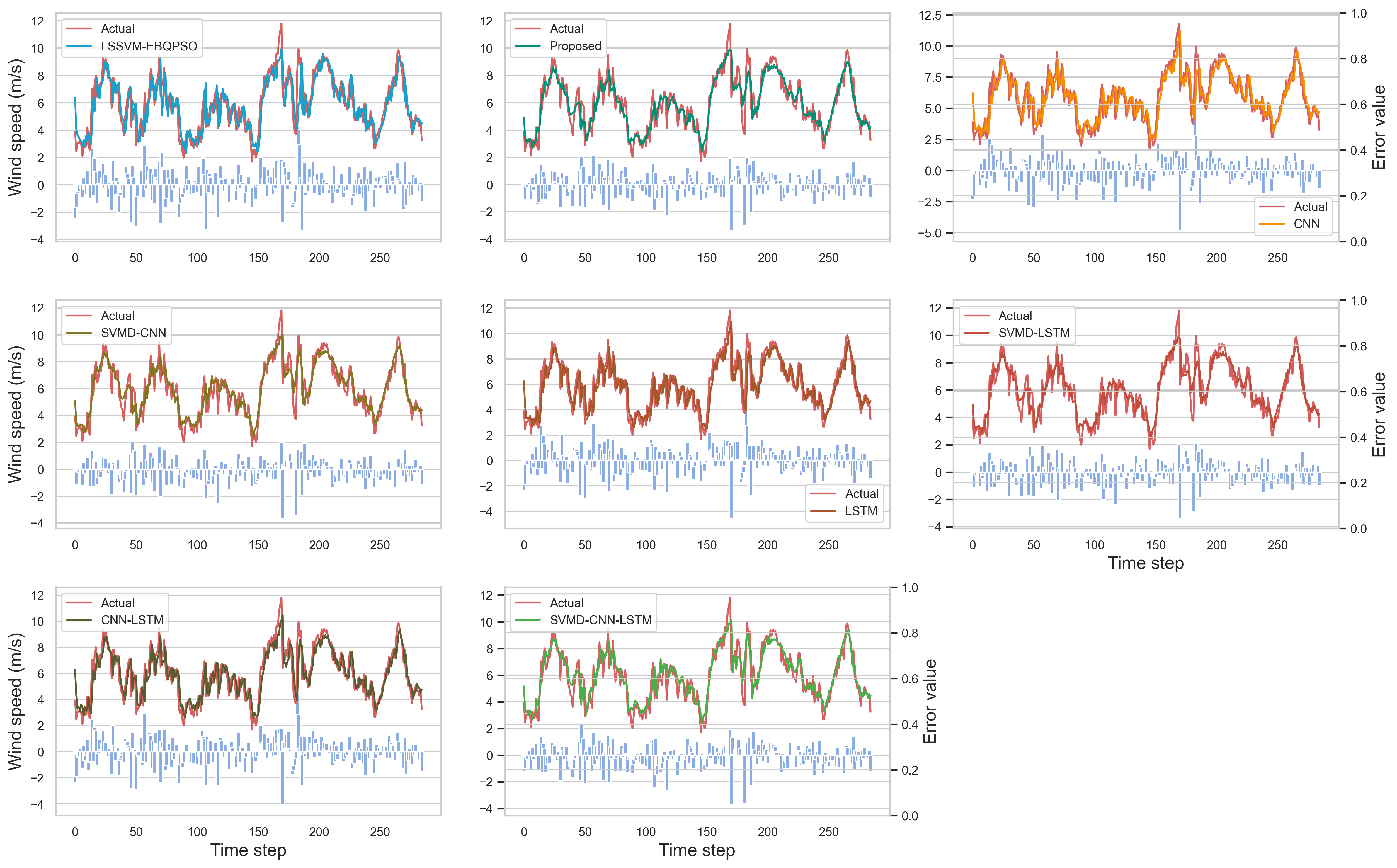}
    \caption{Actual vs Predicted results and the corresponding error bars for the May Dataset.}
    \label{fig:act_pred_2}
\end{figure}

Similarly, for the May dataset, the proposed model generated superior results as compared to the benchmark methods, except for the SVMD-LSTM model, in which case both methods obtained similar results. The proposed method scored 0.856 in RMSE, 0.661 in MAE, 13.0\% in MAPE, 0.817 in R$^2$, and 0.905 in correlation coefficient. The least-performing method is the EBQPSO-LSSVM model. The proposed system attained a 31.66\% RMSE, 32.68\% MAE, 28.46\% MAPE, 19.62\% R$^2$, and 9.17\% correlation coefficient improvement over the EBQPSO-LSSVM model. This gain in performance proves the impact of the SVMD and LSTM methods on the overall model improvement.

As compared to the SVMD-LSTM model, the proposed model scored less than 1\% higher MAPE while obtaining the same performance in terms of RMSE and R$^2$. Further, the SVMD-LSTM method produced less than 1\% performance improvement in MAE and less than 0.2\% improvement in terms of correlation coefficient, which by all accounts is almost negligible. Thus, despite the performance closeness of the SVMD-LSTM model with the proposed method for the May dataset, this similarity vanishes, and the superiority of the proposed method is retained when both datasets are taken into account. On average, the proposed model achieves a 5.76\% RMSE, 8.85\% MAE, and 5.93\% MAPE improvement over SVMD-LSTM. 

To further elucidate the forecasting capabilities of the proposed method, Figures \ref{fig:act_pred_1} and \ref{fig:act_pred_2} show the actual versus the predicted values for the proposed and benchmark functions, along with the error indices for both datasets. The error values are computed by taking the difference between the actual wind speed and the predicted wind series. The fitting capability of the proposed approach is good, as it was able to recognize all the patterns of the test sets (April and May) with great accuracy. Moreover, the benchmark methods also displayed great generalization capability on the test sets. Validating conclusively that the proposed system demonstrated better generalization capabilities by merely observing the graphs is a difficult task. This is due to the size of the wind test set not being large enough to identify nuances that explain the proposed model's superiority. However, it can be observed that the error indices are slightly smaller in magnitude than those of the benchmark models.

\begin{figure}[!ht]
    \centering
    \includegraphics[scale=.45]{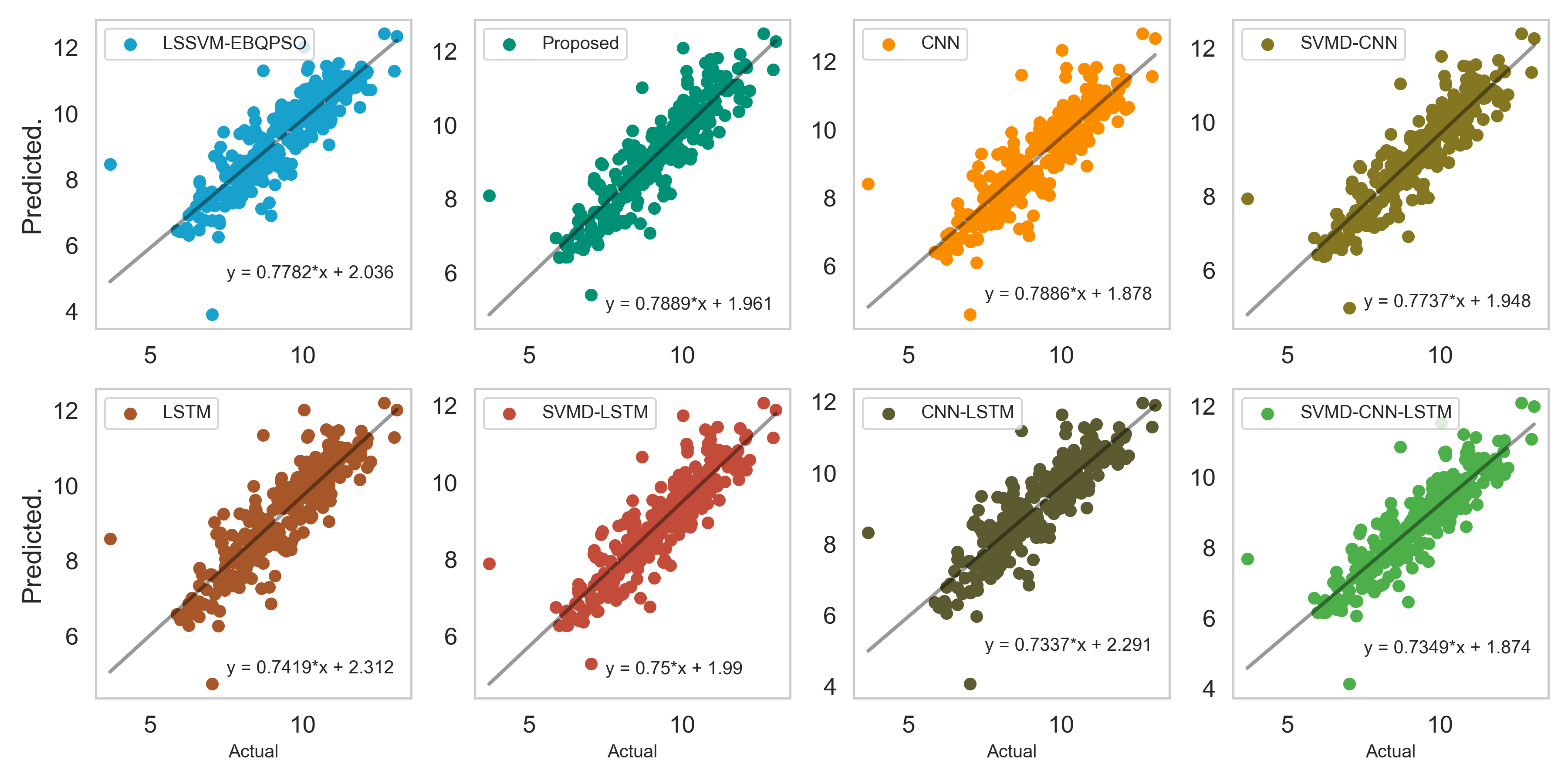}
    \caption{Linear fit of the actual vs predicted wind speed values (April dataset).}
    \label{fig:ds1_linearfit}
\end{figure}

\begin{figure}[!ht]
    \centering
    \includegraphics[scale=.45]{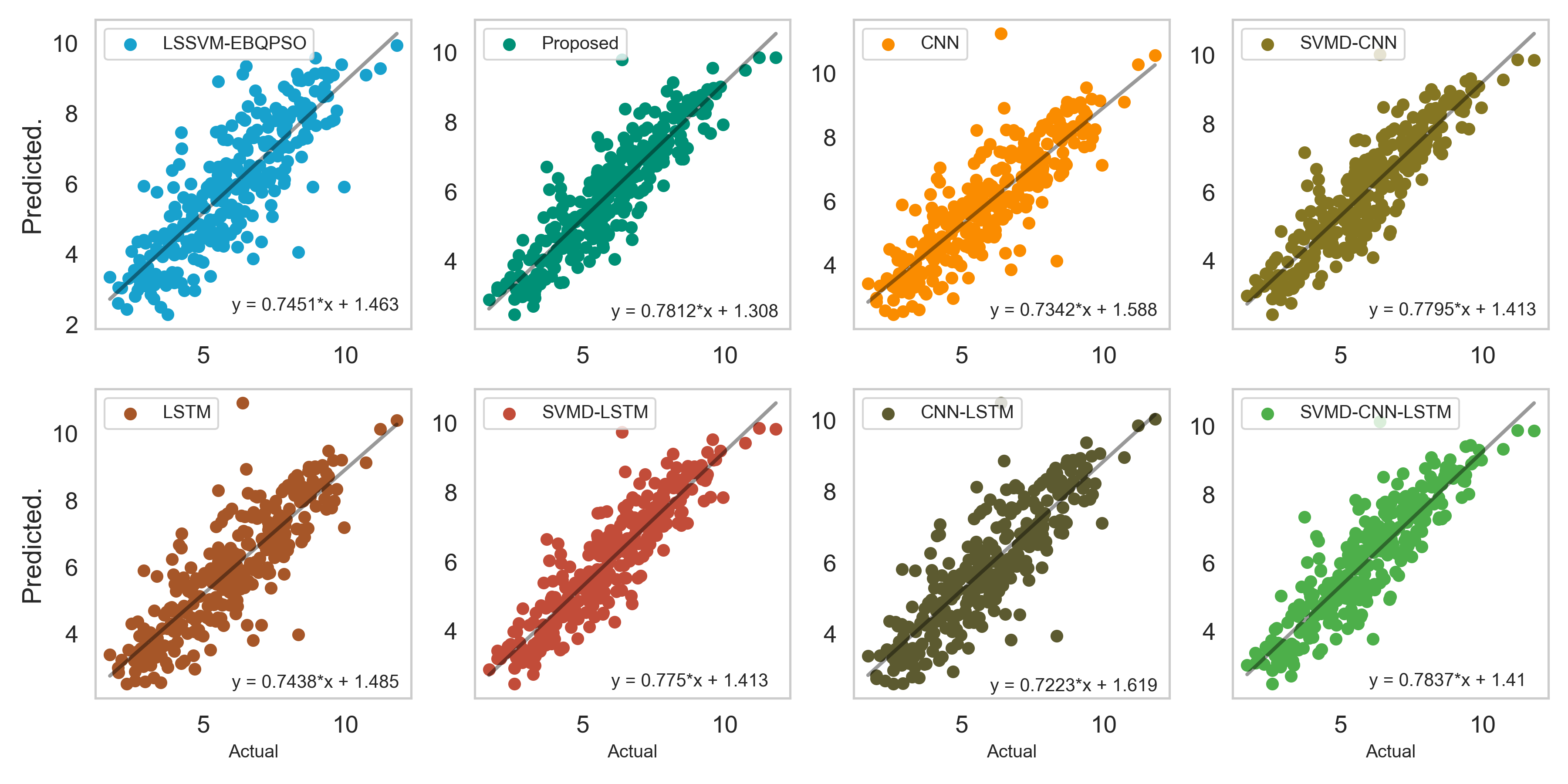}
    \caption{Linear fit of the actual vs predicted wind speed values (May dataset).}
    \label{fig:ds2_linearfit}
\end{figure}

Another important performance assessment tool for forecasting models is the linear fit. An ideal model is one that results in the same predicted value for each actual value. In such a case, the slope of the linear plot is one. In general, a good fit is one that keeps the predicted values in close proximity to the actual values. Therefore, the robustness of the model is inferred from how condensed the points are and the slop of the linear plot. Figures \ref{fig:ds1_linearfit} and \ref{fig:ds2_linearfit} depict the linear fit plots of the proposed method and the benchmark methods for the April and May datasets, respectively, including the actual versus predicted value plots, the linear fit, and the expression of the linear fit, $y \, = \, m*x + b$, where $m$ is the slope of the linear plot and $b$ is the y-intercept.

For the April dataset, we have points of the proposed SVMD-EBQPSO-LSSVM-LSTM model more condensed around the linear plot. Moreover, the slope of the proposed approach's linear plot, 0.7889, is the closest to one. From this, it can be concluded that the proposed approach displayed better forecasting capability than the benchmark methods. Correspondingly, for the May dataset, the proposed method attained the most condensed points to the linear fit and achieved the highest slope, 0.7812, as compared to the benchmark methods. Hence, it can be concluded that the proposed method is the best for fitting both the April and May test sets.

\begin{figure}[!ht]
    \centering
    \includegraphics[scale=.45]{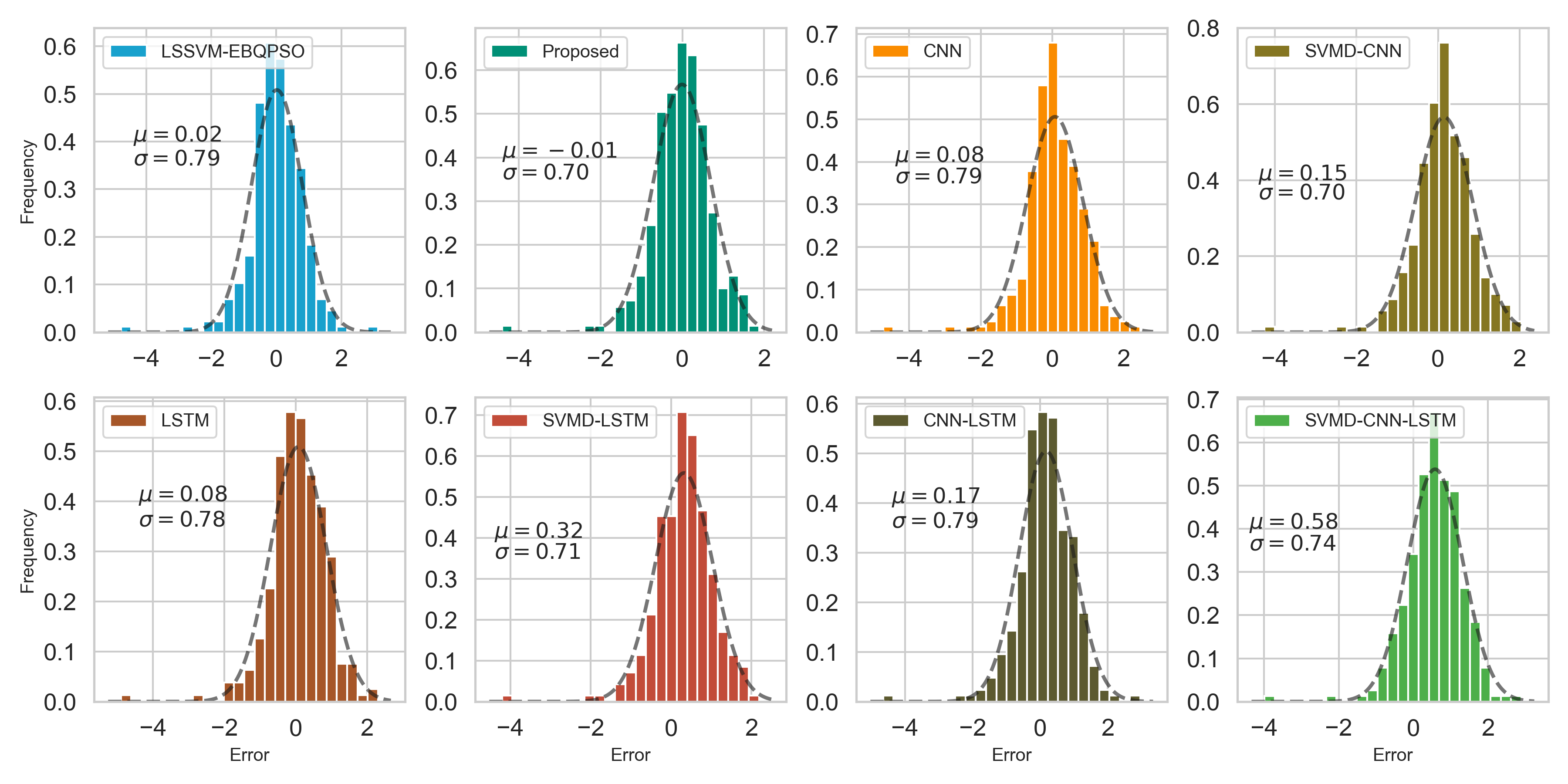}
    \caption{Error distribution plots (April Dataset)}
    \label{fig:ds1_errordist}
\end{figure}

\begin{figure}[!ht]
    \centering
    \includegraphics[scale=.45]{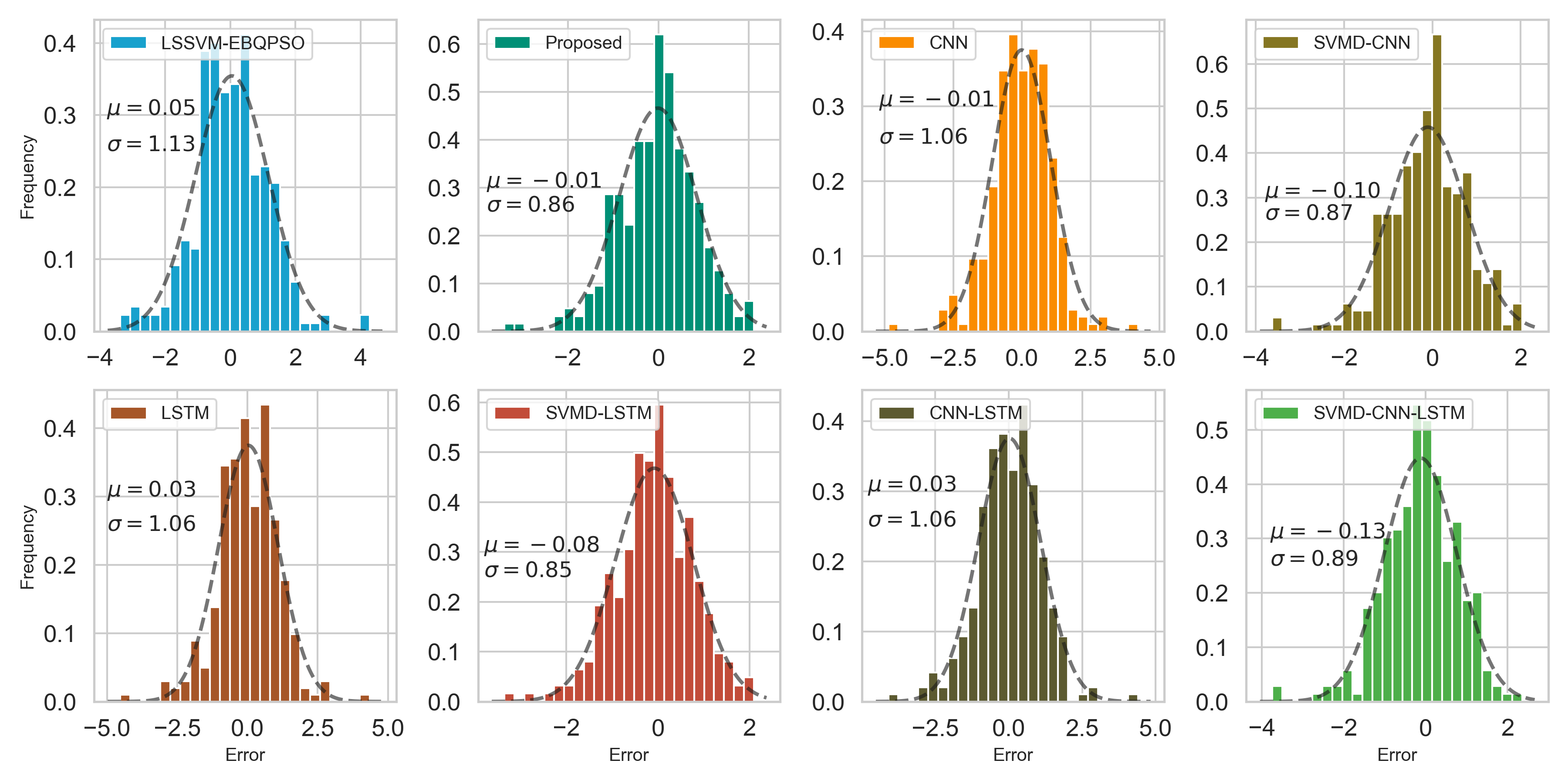}
    \caption{Error distribution plots (May Dataset)}
    \label{fig:ds2_errordist}
\end{figure}

The distribution of the residual errors is also presented in Figures \ref{fig:ds1_errordist} and \ref{fig:ds2_errordist} for the April and May datasets, respectively. It can be observed that for all the models, the distributions can be estimated by a normal distribution with varying mean and variance. In general, a good fit is expected to have a narrow range (that is, a small variance) centered at the center of the distribution (that is, a zero mean). From Figure \ref{fig:ds1_errordist}, the proposed method generated error distributions with the lowest mean ($\mu = -0.01$) and variance ($\sigma = 0.70$). The SVMD-CNN model also has the same variance but a mean higher than the proposed method. Similarly, the proposed method produced residual error distributions with the mean nearest to zero ($\mu = -0.01$) and the lowest variance ($\sigma=0.86$) as compared to the benchmark models. The CNN model is another model that achieved the same mean but a variance larger than that of the proposed approach ($\sigma=1.06$). Therefore, both figures further reinforce the superiority of the proposed model in accurately predicting and generalizing unseen sequences.

In conclusion, the proposed SVMD-EBQPSO-LSSVM-LSTM model has demonstrated superior performance over well-known models as validated using various forecasting metrics and separate datasets.

%
\section{Conclusion and Future Directions}
\label{sec:conclusion}
At present, renewable energy and machine learning are the fastest-growing fields of study due to the alarming need to protect the environment and energy and provide cost-efficient and automated mechanisms to do so. The advent of artificial intelligence and machine learning tools is providing a new paradigm for developing systems that efficiently utilize renewable energy sources. With this in mind, this work proposes a hybrid machine learning model that takes into account the intermittent and non-stationery characteristics of wind speed, the source of wind power generation in wind farms. The proposed SVMD-EBQPSO-LSSVM-LSTM model uses SVMD for decomposing the original wind series into intrinsic mode functions, EBQPSO for tuning the LSSVM hyperparameter, and LSTM for modeling differential errors between the sum of the modal component and the original wind series. By taking two different wind speed data sets collected from a local wind farm and using different performance, the proposed model demonstrated superior overall performance as compared to well-known forecasting benchmark methods.

As a future work, we recommend testing and validating the proposed approach using wind data set of larger sizes with more compute power.


 \section{Data Availability}
\label{sec:availability}
 Authors declare that all data including code implementations and data sets will be available upon request.

  \bibliographystyle{elsarticle-num} 
  \bibliography{references.bib}


\end{document}